\PassOptionsToPackage{table}{xcolor} 
\documentclass{fairmeta}

\usepackage{amsmath}        
\usepackage{amsfonts}       
\usepackage{url}            
\usepackage{nicefrac}       
\usepackage{float}          
\microtypesetup{expansion=false,protrusion=false}

\definecolor{tabGrpA}{HTML}{E8F0F9}
\definecolor{tabGrpB}{HTML}{E6F2EA}
\definecolor{tabGrpC}{HTML}{FBF0DC}

\title{\fontsize{18.8}{22}\selectfont World-R0: A Unified Representation for World Modeling}

\author[*,\dagger]{Yifeng Shi}
\author[*]{Mingwei Xing}
\author[*]{Xinliang Wang}
\author{Jinrang Jia}
\author{Zhenjia Li}
\affiliation{Ke Holdings Inc.}
\contribution[*]{Equal contribution}
\contribution[\dagger]{Corresponding author: \email{shiyifeng@tju.edu.cn}}

\abstract{
World models encompass tasks such as video generation, action generation, and 3D scene generation, each relying on different visual representations. Existing approaches adapt foundation representations separately for specific tasks, making it difficult to establish a unified, reusable interface and potentially losing information from the original representations during adaptation. We propose World-R0, a general-purpose world representation that supports causal updates and can serve as a unified interface across different world models. First, we systematically compare different representation adaptation methods and explore a simple and effective post-training method based on asymmetric distillation. A causal encoder is supervised by a bidirectional video teacher with full spatiotemporal context and an image teacher that processes individual frames independently, enabling the encoder to learn temporal dynamics and fine-grained semantics while accessing only current and past frames. Across five probing tasks with frozen encoders, World-R0 achieves broadly comparable performance to that of bidirectional V-JEPA2.1-L while supporting causal updates and temporal compression. 
Second, we observe that generation quality is associated with how well the noise schedule matches the energy distribution of the representation. By analyzing feature covariance spectra and feature recoverability after noise addition, we propose feature-energy-aware shift modulation to better align the noise schedule with the energy structure of each representation, thereby improving generation quality. Finally, we evaluate World-R0 on three types of world-modeling tasks. For video generation, it achieves gFVD scores of 114.96, 66.07, and 51.51 on UCF101, Diving48, and K600, respectively, outperforming V-RAE under matched DiT settings on all three datasets. For action generation, it achieves an average success rate of 97.75\% on LIBERO with 0.9B parameters. For 3D Gaussian scene generation, it achieves a gFVD score of 93.21 on RE10K without explicit camera pose inputs. These results demonstrate that World-R0's causal representation supports video generation, action generation, and 3D scene generation while preserving its understanding capabilities.

\noindent\textbf{Project page:} 
\url{https://xmw666.github.io/world-r0-pages/}
}

\begin{document}
\maketitle

\begin{figure}[H]
  \centering
  \includegraphics[width=0.88\textwidth]{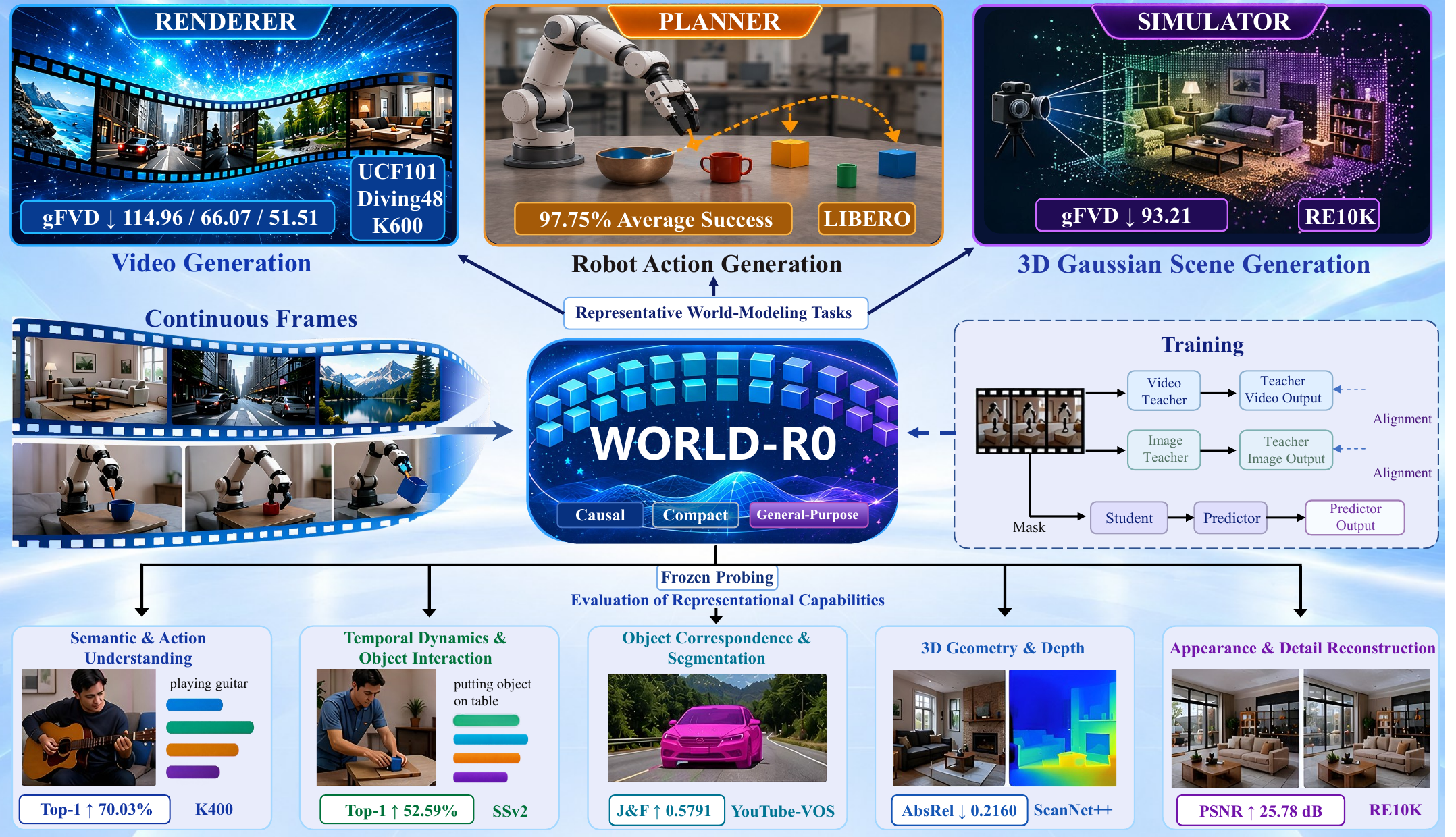}
  \caption{\textbf{World-R0 is a general-purpose world representation that supports causal updates and can serve as a unified interface across different world models.}}
  \label{fig:teaser}
\end{figure}

\section{Introduction}

Recent research on world models can be broadly categorized into three directions, focusing on visual rendering, predictive planning, and environment simulation. The first direction is video generative modeling, which typically encodes visual observations into video latents and generates future videos through diffusion-based or autoregressive modeling~\cite{cogvideox,wan,ltxvideo,hunyuanvideo,opensora}. The second direction is latent-space dynamics prediction, which does not directly generate pixels but instead predicts future states in the visual feature space, providing a basis for scene understanding, planning, and action decision-making~\cite{planet,dreamer,muzero,vjepa2}. The third direction is explicit 3D scene representation and generation, which recovers scene geometry, camera parameters, and appearance from images or videos, and reconstructs or generates explicit environment representations, typically in the form of 3D Gaussians, thereby providing a spatial foundation for environment simulation~\cite{panoworld,adapsplat,artifactworld,pano2world,lyra,yogo,cruise}.

The three directions have different requirements for visual representations. Video generation models typically employ spatiotemporally compressed latents, such as those produced by video VAEs, with CogVideoX~\cite{cogvideox} as a representative example. Some 3D reconstruction and scene generation methods leverage pretrained 3D representations. For example, Gen3R and OneWorld~\cite{gen3r,oneworld} integrate such representations with generative models. Latent-space predictive models directly perform state modeling based on pretrained visual features. For instance, DINO-WM~\cite{dinowm} predicts frozen DINOv2~\cite{dinov2} features, while V-JEPA 2~\cite{vjepa2} predicts future states in the video feature space and supports action-conditioned planning. These advances demonstrate that visual representations play an important role in generation, 3D modeling, and future-state prediction, yet their selection and adaptation remain highly task-dependent.

To leverage existing foundation representations across different world-modeling tasks, adaptation is often required to enable temporal modeling, spatiotemporal compression, and causal updates~\cite{raev2,fae,phoenix,vrae,oneworld,gen3r}. For example, Phoenix~\cite{phoenix} applies a causal semantic autoencoder to temporally aggregate and compress frame-wise visual features into a compact latent that depends only on current and past observations, supporting joint generation of visual states and actions. V-RAE~\cite{vrae} aggregates and compresses features across adjacent temporal positions through learnable temporal attention pooling, jointly trains a causal decoder to reconstruct continuous videos, and trains a DiT in the compressed latent space for video generation.
However, existing representation adaptation methods still face three limitations. First, adaptation is typically task-specific, making the resulting representations difficult to reuse as a shared interface across different world models. Second, representation adaptation may cause information loss. Temporal compression and causal adaptation may alter the information distribution of foundation representations, affecting the retention of semantic, temporal, spatial, and appearance information. Therefore, the evaluation results of the original representations cannot directly reflect the capabilities of the adapted representations. Third, a gap still exists between representational capabilities and suitability for generative modeling. Strong performance on understanding or prediction tasks does not necessarily imply that the underlying feature distribution can be effectively learned by a generative model. Even latents with identical token counts and channel dimensions can differ in their energy distributions and spectral structure. Reusing existing generation settings may therefore fail to realize the full potential of these representations.

To address these problems, we first compare in detail the influence of different representation adaptation methods on representational capabilities, and explore a simple and effective asymmetric distillation post-training method for learning a compact representation that supports causal updates (World-R0). This method takes the bidirectional V-JEPA2.1~\cite{vjepa21} as the teacher, transfers its spatiotemporal modeling capability to the causal encoder, and introduces a single-frame DINOv3~\cite{dinov3} teacher to provide fine-grained semantic and appearance supervision, enhancing the retention of local information. The trained encoder directly processes visual observations, and when new observations arrive, past computations can be reused via the KV cache. We design five probing tasks to evaluate semantics, temporal modeling, spatial structure, and appearance reconstruction. World-R0 reaches 70.03\% and 52.59\% on K400~\cite{kinetics} and SSv2~\cite{ssv2} video action recognition, respectively; on YouTube-VOS~\cite{youtubevos} object segmentation, ScanNet++~\cite{scannetpp} depth estimation, and RE10K~\cite{re10k} appearance reconstruction, it obtains J\&F of 0.5791, AbsRel of 0.2160, and PSNR of 25.78 dB, respectively. Compared with the bidirectional V-JEPA2.1-L~\cite{vjepa21} under the same configuration, World-R0 performs better on video classification and appearance reconstruction, while maintaining comparable performance on object segmentation and depth estimation.

To make the learned representation suitable for generation, we further study the alignment between the feature energy distribution and noise injection. Existing shifts based on latent dimensionality fail to capture differences in energy distributions across representations~\cite{sd3,rae,vrae}. We analyze the relationship between the feature covariance spectrum and the signal-to-noise ratio along each feature direction after noise injection, and propose \textbf{feature-energy-aware shift modulation}, which adjusts the shift using the log-spectral width and the proportion of effective directions. Ablation experiments demonstrate the benefits of this adjustment over the default shift. In video generation experiments on UCF101~\cite{ucf101}, Diving48~\cite{diving48}, and K600~\cite{k600} with 4$\times$ temporal compression, World-R0 achieves gFVD scores of 114.96, 66.07, and 51.51, respectively, improving on V-RAE's~\cite{vrae} corresponding scores of 117.86, 87.96, and 52.26.

Beyond video generation, we also apply the World-R0 representation to WAM action generation and 3D Gaussian scene generation. For action generation, the model performs one-step prediction of future latents based on current and past observations, then decodes these latents into robot-arm action trajectories. Causal encoding supports incremental updates as new observations arrive. The model achieves an average task success rate of 97.75\% on LIBERO~\cite{libero}, exceeding V-JEPA-Policy~\cite{vjepa-policy} by 0.9 percentage points. For 3D Gaussian scene generation, the model generates latents conditioned on the input image, and a Gaussian decoder maps these latents to a 3D scene. On RE10K~\cite{re10k}, it achieves a gFVD score of 93.21 without explicit camera pose inputs. Together with the probing results, these experiments show that the learned representation retains information useful for understanding and that predicted or generated latents can be decoded for different downstream tasks.

Our main contributions are as follows:

\begin{itemize}

\item

\textbf{Exploration of representation learning methods.} We systematically compare the effects of different adaptation methods on representational capabilities, investigate three feasible adaptation strategies, and ultimately identify a simple yet effective post-training scheme for developing World-R0, a compact and general-purpose world representation that supports causal updates. This approach takes the bidirectional V-JEPA as the teacher and the causal encoder as the student, and performs asymmetric distillation combined with DINO fine-grained semantic supervision. World-R0 reaches 70.03\% and 52.59\% on K400 and SSv2, respectively, obtains J\&F of 0.5791 and AbsRel of 0.2160 on YouTube-VOS object segmentation and ScanNet++ depth estimation, respectively, and obtains PSNR of 25.78 dB on RE10K appearance reconstruction; compared with bidirectional V-JEPA2.1-L, World-R0 maintains competitive overall probing performance under causal constraints.

\item

\textbf{Feature-energy-based generation adaptation.} We propose feature-energy-aware shift modulation, which corrects the shift parameter during generation according to the feature covariance spectrum so that the noise schedule better matches the energy structure of the representation, thereby improving the quality of video generation based on it. In video generation experiments on UCF101, Diving48, and K600, it obtains gFVD scores of 114.96, 66.07, and 51.51, respectively, all better than the corresponding results of V-RAE.

\item

\textbf{Evaluation across world-modeling tasks.} We apply World-R0 to video generation, WAM action generation, and 3D Gaussian scene generation to evaluate its applicability to the renderer, planner, and simulator paradigms of world modeling, respectively. On the LIBERO action-generation task, World-R0 achieves an average task success rate of 97.75\%, outperforming V-JEPA-Policy by approximately 0.9 percentage points and demonstrating superior action generation capabilities. It also achieves a gFVD score of 93.21 in 3D generation experiments on RE10K.  Together with the aforementioned representation probing and video generation experiments, these results demonstrate that World-R0 can support multiple world-modeling paradigms with a unified causal representation while preserving its visual understanding capabilities.

\end{itemize}

\section{Related Work}

\subsection{Visual Representation Learning}

Visual foundation representation learning has expanded from single-frame semantic understanding to dense spatial information and video temporal modeling~\cite{clip,vjepa21,dinov3,vitcomer,star}. The DINO series \cite{dino,dinov2,dinov3} learns image features with semantic and spatial structure through unsupervised self-distillation and large-scale training; CLIP and SigLIP learn open-vocabulary semantic representations through image-text alignment \cite{clip,siglip}. MAE, iBOT, and I-JEPA learn image representations through pixel reconstruction, online tokenization, and feature prediction, respectively \cite{mae,ibot,ijepa}.
Video foundation representations introduce cross-frame modeling. The VideoMAE series uses high-ratio tube masking to learn spatiotemporal features, while InternVideo combines masked video modeling with video-text contrastive learning \cite{videomae,videomae2,internvideo}. V-JEPA directly predicts latent spatiotemporal features and avoids using pixel reconstruction as the primary learning objective \cite{vjepa}; V-JEPA 2 and V-JEPA 2.1 further use video features for future-state prediction, action-conditioned modeling, and dense visual understanding \cite{vjepa2,vjepa21}. These methods can model video temporal structure and support multiple downstream tasks, but most use bidirectional encoding, so historical frames must still be reprocessed when new observations arrive. We use a simple and effective asymmetric distillation post-training method to transfer the spatiotemporal prior of bidirectional V-JEPA and the fine-grained semantic and appearance priors of single-frame DINO to a causal encoder, obtaining compact visual features that support causal updates.

\subsection{Visual Representations for Generation}

Mainstream visual generation models usually first construct spatiotemporal latents through a VAE or VQ tokenizer, and then model them with an autoregressive Transformer or diffusion Transformer \cite{videogpt,magvit,magvit2,dit,sit,cogvideox,wan}.
Recent work has begun to use pretrained visual features for generation. In image generation, RAEv2 directly freezes a visual encoder and trains an RGB decoder, whereas FAE adds a low-dimensional bottleneck in visual feature space and reconstructs both features and images separately \cite{raev2,fae}. In video generation, V-RAE uses temporal pooling to compress frozen video foundation representations and trains a video decoder to recover continuous motion \cite{vrae}. REPA and REPA-E use pretrained visual features for intermediate-layer alignment or joint optimization in diffusion models \cite{repa,repae}. In 3D generation, Gen3R and OneWorld map geometric, appearance, and semantic information into unified 3D representation spaces \cite{gen3r,oneworld}. These studies show that existing visual representations can effectively serve generative modeling, but how to improve their suitability for generative modeling while preserving their original representational capabilities warrants further investigation. Existing dimension-dependent shift strategies~\cite{sd3,rae} do not explicitly account for differences in feature energy distributions across representations. We further analyze the relationship between the feature covariance spectrum and noise injection, and propose feature-energy-aware shift modulation to better align the noise schedule with the energy structure of the representation.

\subsection{Visual Representations in World Models}

Visual representations in world models are mainly used for future-state prediction, interactive video generation, action decision-making, and 3D scene modeling. DINO-WM freezes DINOv2, uses visual patch features as the environment state, and predicts future features for goal-conditioned planning \cite{dinowm}; V-JEPA 2 models future states in video feature space and supports robot planning through action-conditioned post-training \cite{vjepa2}.
Another line of work uses video generation models to simulate interactive environments by generating subsequent videos conditioned on historical video frames and actions \cite{genie,videoworld,vid2world,gamengen,diamond,ref39}. Related studies further introduce frame-by-frame generation, long-term memory, and causal attention to improve prediction stability \cite{diffusionforcing,longcontextssm,ref39}.
In contrast, Gen3R and OneWorld target 3D scene generation and map geometric and appearance information into specific 3D representation spaces \cite{gen3r,oneworld}. These methods usually adapt visual foundation representations according to the particular task and model architecture. The adaptation module is often tied to the downstream task and decoder, which makes the resulting adapted representation difficult to reuse directly for other world-modeling tasks. The adaptation process may also change the information distribution of the original foundation representation and lose semantic, temporal, or appearance information. Our visual representation supports causal incremental encoding of continuous observations and can be used across video understanding, reconstruction, video generation, action generation, and 3D tasks.

\section{Temporal Compression and Causal Adaptation}

Temporal compression and causal adaptation are important directions in representation adaptation. They reduce redundancy in continuous observations and enable reuse of past computation, making pretrained visual foundation representations more suitable for generation, prediction, and decision-making under continuous observations. However, these adaptations may affect the original capabilities, so the retention of semantic, temporal, spatial, and appearance information must be evaluated. We therefore establish a frozen probing protocol covering five tasks to provide a unified basis for evaluating different representations and adaptation methods.

We then analyze the limitations of existing foundation and adapted representations across semantic, temporal, spatial, and appearance dimensions based on the evaluation results. Building on this analysis, we further explore three representation learning approaches: Temporal Feature Autoencoder, Feature Prediction Adapter, and Asymmetric Distillation Post-training, systematically comparing their effects on representational capabilities. Sections~\ref{sec:shift-modulation} and~\ref{sec:world-model-tasks} then explore adaptation of representations for generative modeling and applications to other world-modeling tasks, respectively.

\subsection{Evaluation Tasks and Probing Protocol}

\subsubsection{Evaluation Dimensions and Dataset Selection}

Drawing on prior evaluation practices for visual foundation representations~\cite{vjepa21,dinov3,yue2024improving,weijler2026latent}, we select Kinetics-400 (K400), Something-Something V2 (SSv2), YouTube-VOS (YOS), ScanNet++ (SPP), and RealEstate10K (RE10K) to construct five probing tasks that measure information retention after adaptation from complementary perspectives.

K400 primarily evaluates action recognition and scene-level semantics, relying more heavily on single-frame visual content and placing relatively limited demands on cross-frame temporal modeling. SSv2 emphasizes object interactions and temporal changes and therefore imposes stronger requirements on video temporal understanding. YouTube-VOS evaluates object spatial structure and cross-frame correspondence through video object segmentation conditioned on the first frame. ScanNet++ evaluates scene geometry through depth estimation. RE10K evaluates the recoverability of colors, textures, and local details through appearance reconstruction from an input video. The tasks and metrics are listed in Table~\ref{tab:probe-tasks}.

\begin{table}[t]
  \caption{\textbf{Five representation probing tasks and evaluation metrics.} Classification accuracy is reported as a percentage, J\&F ranges from 0 to 1, and PSNR is measured in dB.}
  \label{tab:probe-tasks}
  \centering
  \resizebox{\textwidth}{!}{
  \begin{tabular}{lllll}
    \toprule
    Dataset & Probing task & Training set & Evaluation set & Metric \\
    \midrule
    K400~\cite{kinetics} & Video action recognition & 240,436 videos & 19,796 videos & Top-1 accuracy $\uparrow$ \\
    SSv2~\cite{ssv2} & Video action recognition & 168,913 & 24,777 & Top-1 accuracy $\uparrow$ \\
    YouTube-VOS~\cite{youtubevos} & Video object segmentation & 3,471 videos & 507 videos & J\&F $\uparrow$ \\
    ScanNet++~\cite{scannetpp} & Video depth estimation & 856 scenes & 50 scenes & AbsRel $\downarrow$ \\
    RE10K~\cite{re10k} & Video appearance reconstruction & 58,515 scenes & 7,286 scenes & PSNR $\uparrow$ \\
    \bottomrule
  \end{tabular}
  }
\end{table}
Here, J\&F combines region similarity J and contour accuracy F; AbsRel measures the absolute relative error of predicted depth with respect to ground-truth depth; and PSNR quantifies pixel-wise reconstruction fidelity between the reconstructed and reference images. Together, these five tasks assess representational capabilities across video understanding, dense prediction, and appearance reconstruction, providing a unified basis for comparing the adaptation methods discussed below.

\subsubsection{Probing Head Design and Training Setup}

To assess the effects of different adaptation methods while controlling for differences in datasets and network architectures, we freeze the representation encoder and train only the probing heads. The task-specific probing-head structures and training objectives are as follows. Detailed experimental parameters and optimization settings are given in the experiments section.

\textbf{Video action recognition.} The K400 classification head consists of LayerNorm followed by a linear classifier that predicts one of 400 action classes. It is trained with cross-entropy loss. The SSv2 classification head consists of LayerNorm and a two-layer MLP with GELU activation and predicts 174 action classes; the MLP consists of a hidden layer with 1,024 units, followed by an output layer with 174 units. SSv2 also uses cross-entropy loss. Both tasks report validation Top-1 accuracy.

\textbf{Video object segmentation.} For YouTube-VOS, we initialize object representations using the first frame with foreground annotations and its corresponding masks. The decoder then predicts object masks from subsequent video features. The frozen features are temporally upsampled before entering the segmentation decoder; both the temporal module and decoder have hidden dimension 256. Training computes Dice and binary cross-entropy losses on valid objects and annotated frames, excluding the conditioning frame from the loss. During evaluation, J and F are computed for each object in a sequence and then averaged.

\textbf{Video depth estimation.} The ScanNet++ depth head consists of LayerNorm, two 3$\times$3 convolutions, and a 1$\times$1 convolution that outputs single-channel depth, followed by bilinear interpolation to recover the spatial resolution. For temporally compressed features, a corresponding temporal upsampling module is used according to the compression factor to restore the original temporal resolution, where the stride of the transposed convolution matches the temporal compression factor, with hidden dimension 256. L1 loss is computed only on valid depth pixels. The data are split into training and validation sets by scene. Evaluation covers all videos in the validation scenes, uses a 16-frame sliding window, removes duplicate overlapping frames, and reports AbsRel over the complete videos.

\textbf{Video appearance reconstruction.} RE10K uses a progressive RGB decoding head to recover the input video from frozen features. For temporally compressed features, a corresponding temporal upsampling module is used according to the compression factor to restore the original temporal resolution, where the stride of the transposed convolution matches the temporal compression factor, and the feature widths of both the temporal module and RGB decoder are set to 512. The supervision target is the input RGB image with values in [0,1]. The training loss combines L1 reconstruction loss with a VGG-LPIPS perceptual loss weighted by 0.2. Training and evaluation use the official metadata split. For each scene, we select 16 consecutive frames, resize them to cover the target dimensions, and crop to the input size. We report reconstruction PSNR.

\subsection{Representation Adaptation}

\begin{figure}[t]
  \centering
  \includegraphics[width=\textwidth]{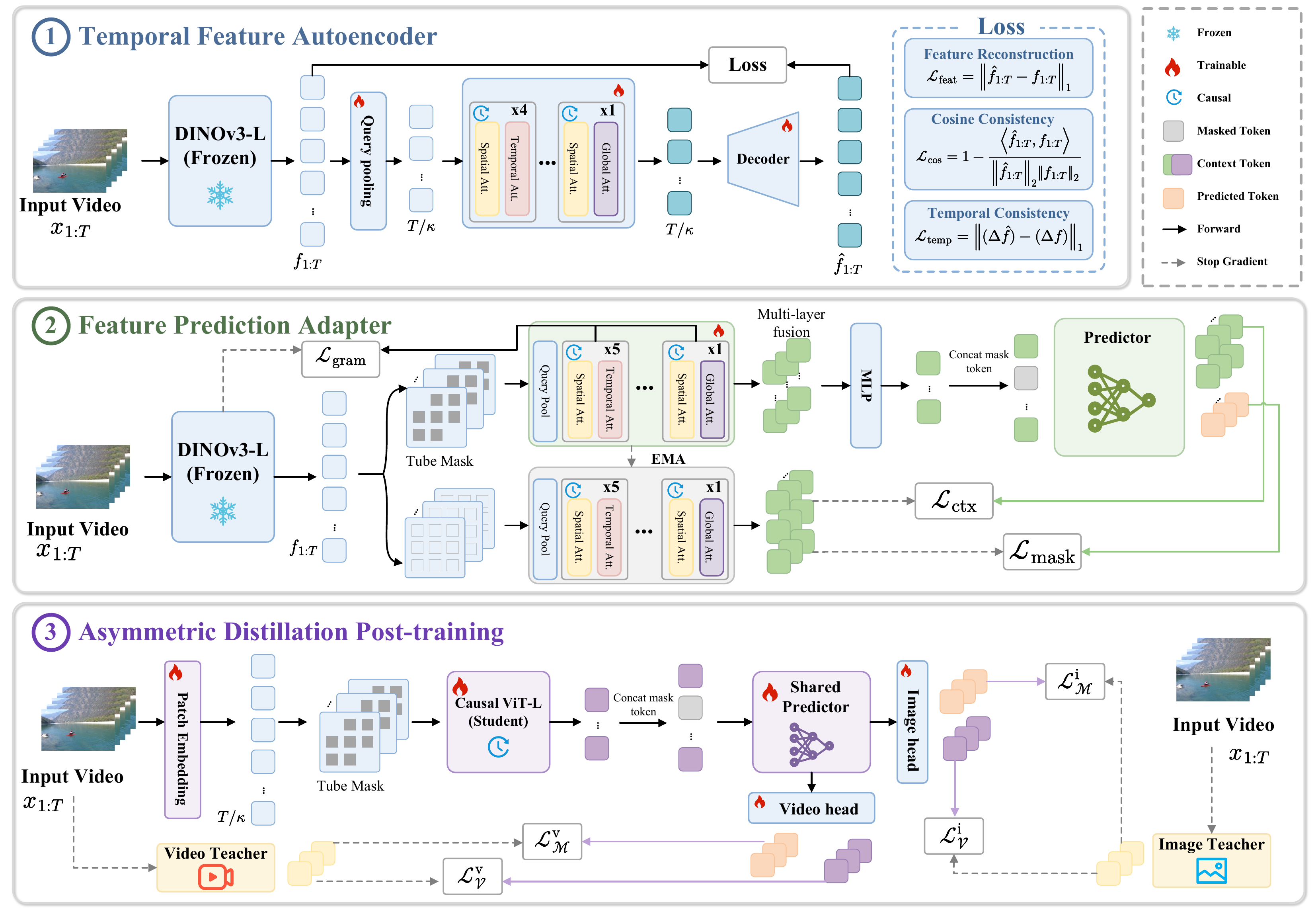}
  \caption{\textbf{Comparison framework for the three representation adaptation methods explored in this paper.}}
  \label{fig:framework}
\end{figure}

Existing representation adaptation methods usually aim to improve temporal compression and causal updates, but our observations show that inappropriate adaptation can degrade the representational capabilities of the original encoder. For example, V-RAE~\cite{vrae} shows lower performance than the corresponding pretrained representation V-JEPA2.1~\cite{vjepa21} on most probing tasks. We therefore compare existing adaptation approaches and introduce three additional methods (Figure~\ref{fig:framework}), including asymmetric distillation post-training, which performs best in validation. Their detailed comparisons are presented in the experimental section.

All three methods target compact spatiotemporal representations that support causal updates. Given a video observation sequence $x_{1:T}$, the encoder $E_\theta$ maps it to

\begin{equation}
z=E_\theta(x_{1:T})\in\mathbb{R}^{T'\times H'\times W'\times C},
\qquad T'=T/\kappa,
\end{equation}

where $T$ is the number of input frames, $\kappa$ is the temporal compression factor, $H'\times W'$ is the spatial feature resolution, and $C$ is the number of channels. We use 2$\times$ temporal compression by default and compare 4$\times$ compression in the ablation experiments.

Temporal causality is imposed on the compressed temporal units. Each temporal unit can access only the current and historical units, not future observations. Therefore, when the temporal partition and positional encoding remain unchanged, appending new observations does not change the representations of existing temporal units, and their attention keys and values can be stored and reused through a KV cache. This property enables incremental updates under continuous observations.

\subsubsection{Temporal Feature Autoencoder}

Given an input video, following the multi-layer feature extraction and fusion approach of RAEv2, DINOv3-L extracts K7 multi-layer features frame by frame and fuses them into a feature sequence $f_{1:T}$ with 1,024 channels. The causal temporal encoder aggregates consecutive frames within a local temporal window through query pooling and compresses the temporal length from $T$ to $T/\kappa$. On the temporally compressed latent sequence, the encoder stacks four spatiotemporally decoupled causal-attention blocks and one causal full-attention block. The first four layers model spatial relations within each temporal unit and temporal relations between the current and historical units at corresponding spatial locations; the causal mask prevents features from accessing future observations. The final layer applies joint attention to all spatiotemporal tokens and integrates a broader spatial and temporal context under the causal constraint. The feature decoder restores the original temporal resolution through temporal upsampling and reconstructs features $\hat{f}_{1:T}$ in the DINOv3 feature space.

The training objective consists of feature reconstruction, feature-direction consistency, and temporal consistency:

\begin{equation}
\mathcal{L}_{\mathrm{AE}}
=\alpha_{\mathrm{feat}}\mathcal{L}_{\mathrm{feat}}
+\alpha_{\mathrm{cos}}\mathcal{L}_{\mathrm{cos}}
+\alpha_{\mathrm{temp}}\mathcal{L}_{\mathrm{temp}}.
\end{equation}
Here, $\mathcal{L}_{\mathrm{feat}}$ constrains the element-wise difference between reconstructed and original features:

\begin{equation}
\mathcal{L}_{\mathrm{feat}}
=\left\|\hat f_{1:T}-f_{1:T}\right\|_1.
\end{equation}

$\mathcal{L}_{\mathrm{cos}}$ constrains the directional consistency between reconstructed and original features:

\begin{equation}
\mathcal{L}_{\mathrm{cos}}
=1-\frac{\left\langle\hat f_{1:T},f_{1:T}\right\rangle}
{\left\|\hat f_{1:T}\right\|_2\left\|f_{1:T}\right\|_2}.
\end{equation}

The temporal consistency loss constrains feature differences at adjacent temporal positions:

\begin{equation}
  \mathcal{L}_{\mathrm{temp}}
  =\left\|(\hat f_t-\hat f_{t-1})-(f_t-f_{t-1})\right\|_1.
  \end{equation}

This constraint makes the reconstruction match not only the content of each frame but also the state changes between adjacent frames. During training, the DINOv3 parameters remain frozen, and only the causal temporal encoder and feature decoder are updated. For downstream use, the input sequence passes through DINOv3 and the causal temporal encoder to obtain the representation.

\subsubsection{Feature Prediction Adapter}

Feature prediction adaptation uses the same frozen DINOv3-L interface as the previous approach but replaces input feature reconstruction with latent feature prediction at masked spatiotemporal positions. Following the V-JEPA training strategy, the model predicts features at masked positions from the visible context to better capture cross-frame temporal relationships.

The model also uses frozen DINOv3-L with K7 multi-layer feature fusion. The input features are temporally compressed by $\kappa\times$ into $T/\kappa$ temporal units. The causal temporal encoder consists of six layers. The first five use point-level spatiotemporally decoupled attention, and each layer performs spatial attention within frames and temporal attention between corresponding spatial locations across frames. The final layer uses causal full spatiotemporal attention to jointly model all spatiotemporal tokens in addition to within-frame attention. Both modules impose a causal constraint along time, so the model first performs temporal interactions at corresponding positions and then integrates a broader spatiotemporal context in the final layer.

During training, we apply a tube mask to the input so that the same spatial region is masked at consecutive temporal positions. The online encoder produces a contextual representation from visible latents, and a predictor then predicts the latents at masked positions. The target branch uses an exponential-moving-average version of the online encoder, namely the EMA target encoder, which encodes the complete feature sequence and provides prediction targets without gradients. The EMA momentum is set to 0.99925.
We use multi-scale supervision. The online encoder extracts intermediate features from layers $1$, $4$, and $6$ and fuses them with an MLP. The predictor takes the fused features as input and predicts the target latents from layers $1$, $4$, and $6$, respectively. Meanwhile, the EMA target encoder independently extracts the corresponding target features from the complete feature sequence, with gradients disabled. Losses between the student predictions and target features are then computed separately at masked and visible positions.

Let $\mathcal{M}$ denote the set of masked positions and $\mathcal{V}$ the set of visible positions. Let $\hat y$ and $y$ denote the predicted and target features, respectively.
The masked-prediction and context-consistency losses are:

\begin{equation}
\mathcal{L}_{\mathrm{mask}} = \sum_{s\in\{1,4,6\}} \lambda_s \left\| \hat y_{\mathcal{M}}^{s} - \operatorname{sg}\left(y_{\mathcal{M}}^{s}\right) \right\|_1,
\end{equation}

\begin{equation}
\mathcal{L}_{\mathrm{ctx}} = \sum_{s\in\{1,4,6\}} \lambda_s \left\| \hat y_{\mathcal{V}}^{s} - \operatorname{sg}\left(y_{\mathcal{V}}^{s}\right) \right\|_1,
\end{equation}

where $s$ denotes the layer and $\operatorname{sg}(\cdot)$ denotes stop-gradient. The loss at masked positions supervises the student to predict masked features from context, whereas the loss at visible positions constrains representations in observed regions to remain consistent with the target encoder and prevents the model from optimizing only masked regions while damaging contextual information.

To preserve token-level relations in the frozen DINOv3 feature space, we further introduce Gram supervision. Let $F_\tau$ and $D_\tau$ denote student features and frozen DINOv3 teacher features at the same temporal position and over the same set of spatial tokens, with each row corresponding to one spatial token. We construct their spatial relation matrices as follows:

\begin{equation}
G_\tau^{\mathrm{s}}=F_\tau F_\tau^{\top},
\qquad
G_\tau^{\mathrm{d}}=D_\tau D_\tau^{\top}.
\end{equation}

The Gram loss matches the relation matrices of the student and teacher and constrains adapted features to preserve spatial relations between tokens. This supervision is applied at layers 1, 3, and 5 of the temporal encoder, and the losses from different layers are summed with weights. The overall training objective is

\begin{equation}
\mathcal{L}_{\mathrm{pred}}
=\alpha_{\mathrm{mask}}\mathcal{L}_{\mathrm{mask}}
+\alpha_{\mathrm{ctx}}\mathcal{L}_{\mathrm{ctx}}
+\alpha_{\mathrm{gram}}\mathcal{L}_{\mathrm{gram}}.
\end{equation}

At inference, no mask is applied and the predictor is not used. The complete video sequence passes through frozen DINOv3 and the EMA causal temporal encoder to obtain representations for downstream probing.

\subsubsection{Asymmetric Distillation Post-Training}

The first two approaches still depend on frozen DINOv3 at inference, and their  representational capabilities are limited by the fixed frame-level interface. Asymmetric distillation post-training directly trains a complete causal video backbone, transferring the dynamic prior from the video teacher and the semantic and appearance priors from the image teacher to the same student network, thereby learning World-R0, a compact and unified visual representation that supports causal updates.

The student network uses causal ViT-L and is initialized with pretrained V-JEPA2.1-L weights. The student performs $\kappa\times$ temporal compression through patch embedding. The frozen video teacher is bidirectional V-JEPA2.1-g and provides cross-frame relations, motion, and dynamic information; the frozen image teacher is DINOv3-H and provides fine-grained semantics and appearance. The asymmetry comes from the different temporal contexts accessible to teacher and student: the video teacher uses bidirectional context from the complete clip to form supervision, whereas the student learns the corresponding spatiotemporal representation under a causal constraint.

During training, the student branch applies a tube mask to the spatiotemporal features after patch embedding and then feeds them to the causal encoder for representation learning; the teacher branches use no mask. The output of the student causal encoder is fed into a shared predictor, whose predictions are then sent to separate video and image prediction heads to map them into the corresponding teacher feature spaces. Additional temporal upsampling modules align the student predictions with the temporal resolution of each teacher, enabling alignment losses to be computed at corresponding spatiotemporal positions in both branches.

Let $q\in\{\mathrm{v},\mathrm{i}\}$ index the video and image teacher branches, respectively, and let the position set be $\mathcal{A}\in\{\mathcal{M},\mathcal{V}\}$. For either branch and position set, the alignment loss is:

\begin{equation}
\mathcal{L}_{\mathcal{A}}^{q}
=\left\|\hat y_{\mathcal{A}}^{q}
-\operatorname{sg}(y_{\mathcal{A}}^{q})\right\|_1,
\end{equation}

where $\hat y_{\mathcal{A}}^{q}$ is the output of the corresponding prediction head and $y_{\mathcal{A}}^{q}$ is the teacher target. The overall distillation objective combines supervision from both teachers at masked and visible positions:

\begin{equation}
\mathcal{L}_{\mathrm{distill}}
=
\alpha_{\mathrm{v}}\mathcal{L}_{\mathcal{M}}^{\mathrm{v}}
+\beta_{\mathrm{v}}\mathcal{L}_{\mathcal{V}}^{\mathrm{v}}
+\alpha_{\mathrm{i}}\mathcal{L}_{\mathcal{M}}^{\mathrm{i}}
+\beta_{\mathrm{i}}\mathcal{L}_{\mathcal{V}}^{\mathrm{i}}.
\end{equation}

The loss at masked positions requires the student to predict teacher features from visible context, whereas the loss at visible positions constrains representation consistency in observed regions. The video teacher provides cross-frame dynamics and event relations, and the image teacher supplements local semantics and appearance details. Both teachers remain frozen during training. The student encoder, shared predictor, and two prediction heads are updated. An EMA of the student parameters is maintained, and the EMA student is used as the encoder for downstream probing.

\subsubsection{Comparison of the Three Methods}

The three representation adaptation methods differ in learning objectives, supervision signals, and temporal modeling approaches. Temporal feature autoencoder adaptation preserves frame-level semantic and appearance information by reconstructing input features, but has limited cross-frame modeling capability. The feature prediction adapter learns spatiotemporal relations by recovering masked latents and uses Gram supervision to preserve spatial structure, but remains limited by the frozen DINOv3 frame-level feature space. Temporal information can only be added to fixed single-frame representations, making it difficult to reorganize low-level features for causal video modeling. Asymmetric distillation post-training uses both video and image teachers: the former provides cross-frame dynamics and spatial structure, while the latter supplements frame-level semantics and appearance information. The causal student aligns with both teacher representations from current and historical context and receives more complete spatiotemporal supervision on top of masked prediction. Although this objective provides weaker direct supervision for features needed for frame-level classification, it achieves a better balance in retaining semantic, temporal, spatial, and appearance information under causal constraints. We therefore adopt this approach to develop World-R0. Detailed quantitative comparisons are presented in Section~\ref{sec:ablation}.

\section{Feature-Energy-Aware Shift Modulation}
\label{sec:shift-modulation}

Existing shift parameters based on latent size do not explicitly account for differences in feature-energy distribution across representations~\cite{sd3,rae}. Different representation learning and adaptation schemes may change the feature energy distribution and channel correlations, thereby forming different covariance spectral structures. To investigate the relationship between these differences and recoverability under noise injection, we first compare the covariance spectra of different representations, then evaluate their reconstruction errors under the same actual noise intensity, and design a feature-energy-aware shift correction method based on the above analysis.

\subsection{Spectral Distributions of Different Representations}

Our analysis uses 2,048 UCF101 videos. We use 70\% to estimate normalization parameters and channel covariance, and reserve the remaining 30\% for reconstruction experiments. Each representation is standardized channel-wise so that its mean feature variance is close to 1, removing differences in overall amplitude. We compare features from the original V-JEPA2.1, V-RAE, and World-R0.

After standardization, each token is treated as a $C$-dimensional vector, and its channel covariance matrix is computed as:

\begin{equation}
\Sigma = \frac{1}{M}\sum_{m=1}^{M}z_mz_m^\top
=U\Lambda U^\top,
\end{equation}

where $z_m\in\mathbb{R}^{C}$ is the channel-standardized and centered token feature vector for token $m$, $C$ is the number of feature channels, and $M$ is the total number of tokens used for covariance statistics. $\Lambda=\operatorname{diag}(\lambda_1,\ldots,\lambda_C)$ is the diagonal matrix of eigenvalues. The $i$-th column of $U$ is a principal direction, and $\lambda_i$ is the variance of features along that direction. In other words, after projecting features onto the $i$-th principal direction, the projection variance is $\lambda_i$. A larger $\lambda_i$ therefore indicates higher feature energy in that direction.

\begin{figure}[t]
  \centering
  \includegraphics[width=\textwidth]{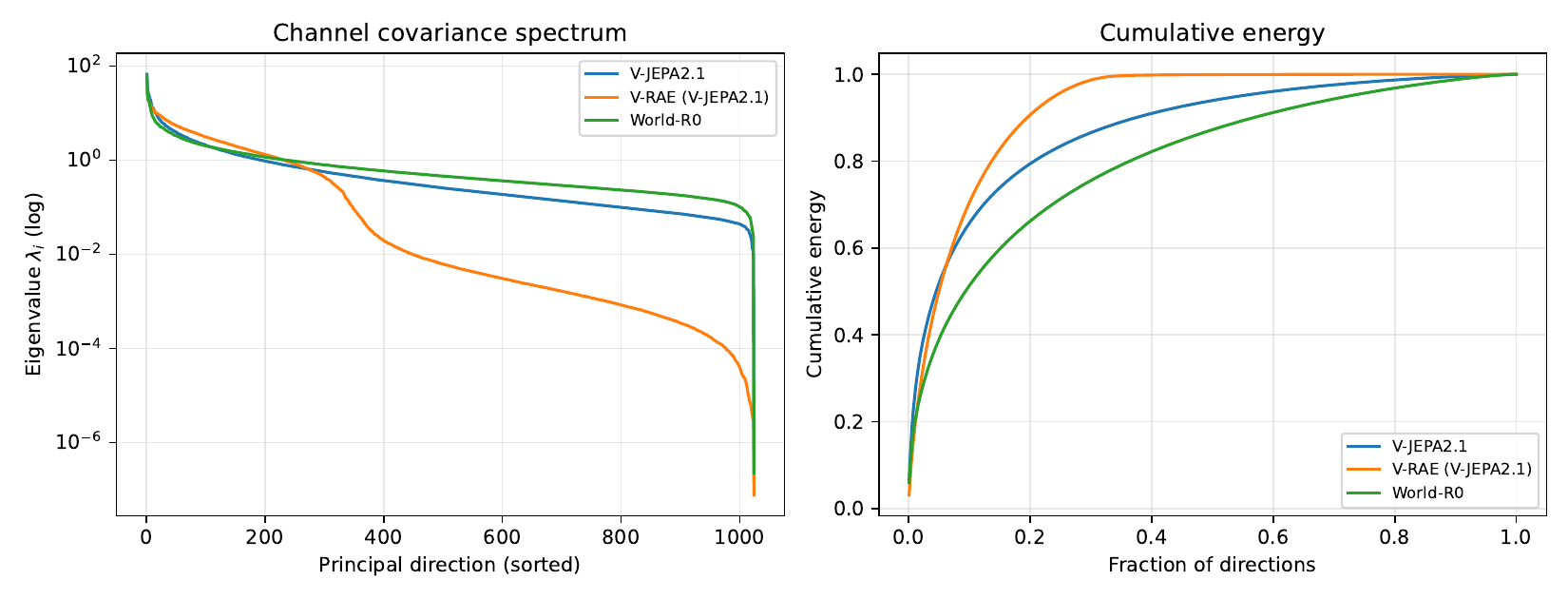}
  \caption{\textbf{Feature covariance spectra and cumulative energy distributions of V-JEPA2.1, V-RAE, and World-R0.} The left panel shows the eigenvalue spectra of the representations, and the right panel shows cumulative energy ratios after sorting eigenvalues in descending order.}
  \label{fig:spectrum}
\end{figure}

The left panel of Figure~\ref{fig:spectrum} shows that the three eigenvalue spectra have distinct shapes. The V-RAE spectrum is dominated by a few high-energy principal directions, followed by a sharp decay and a long low-energy tail. Its energy is therefore concentrated in a small number of principal directions, while many other directions carry almost none. The right panel shows that the top 30\% of V-RAE directions already account for most of the energy. The spectrum of original V-JEPA2.1-L lies between those of V-RAE and World-R0, suggesting that representation adaptation can alter the distribution of feature energy.

\subsection{Noise Sensitivity of Different Representations}

Spectral structure also affects how well a feature can be recovered after noise is added. To verify how recoverability depends on the representation itself and on the noise level, we fix the actual noise level $\tau$ and perturb clean features directly:

\begin{equation}
z_\tau=(1-\tau)z_0+\tau\varepsilon,
\qquad
\varepsilon\sim\mathcal{N}(0,I).
\end{equation}

In this expression, $z_0$ is a standardized clean feature vector, $z_\tau$ is the noisy observation, and $\tau\in[0,1]$ controls the noise intensity. Along the $i$-th direction, the clean-feature variance is $\lambda_i$, so the signal-to-noise ratio is:

\begin{equation}
\operatorname{SNR}_i(\tau)
=
\lambda_i\frac{(1-\tau)^2}{\tau^2}.
\end{equation}

We add noise to held-out samples that were not used to estimate the covariance and recover the original features using a linear estimator derived from that covariance~\cite{kay1993fundamentals}:

\begin{equation}
\hat z_i
=
\frac{(1-\tau)\lambda_i}
{(1-\tau)^2\lambda_i+\tau^2}
z_{\tau,i}.
\end{equation}

The feature recovery error is defined as:

\begin{equation}
\mathcal{E}(\tau)
=
\frac{
\mathbb{E}\|\hat z-z_0\|_2^2
}{
\mathbb{E}\|z_0\|_2^2
}.
\end{equation}

Table~\ref{tab:recovery-error} reports the feature recovery error at several fixed noise levels. 
The recovery error increases with $\tau$ for all representations, with World-R0 exhibiting higher errors. At the same noise level, V-RAE can rely on a few high-energy principal directions that retain relatively strong signals. World-R0 distributes its feature energy across more principal directions with moderate or low energy, resulting in higher normalized linear recovery errors under the same noise levels.

\begin{figure}[t]
  \centering
  \includegraphics[width=\textwidth]{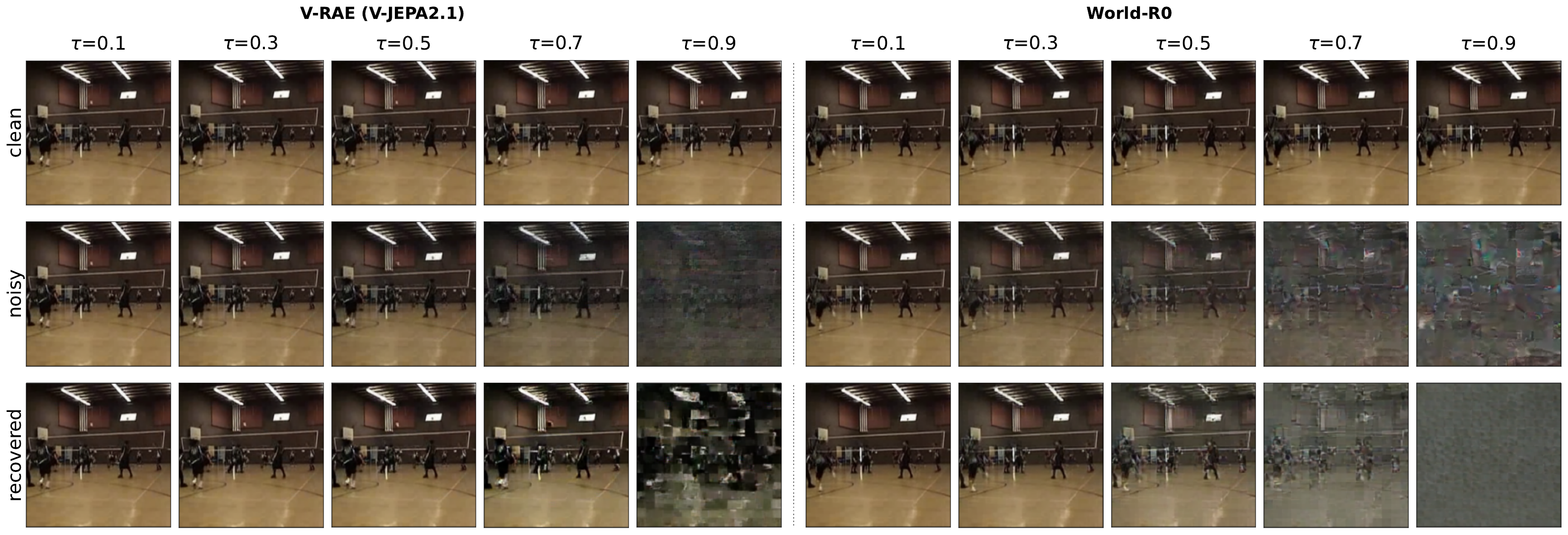}
  \caption{\textbf{Visualization of feature recovery for V-RAE and World-R0 at different noise intensities.} The features recovered by the linear estimator are fed to the corresponding decoder to obtain RGB results. As noise increases, local structures and appearance details in World-R0 exhibit more pronounced degradation, whereas V-RAE retains relatively clear global structure.}
  \label{fig:recovery-vis}
\end{figure}

Figure~\ref{fig:recovery-vis} visualizes reconstruction under different noise levels. At $\tau=0.5$, the corresponding decoder can still recover a fairly clear appearance from V-RAE features, whereas the World-R0 result is visibly blurred and its local details are difficult to distinguish. The qualitative results agree with Table~\ref{tab:recovery-error} and confirm that spectral structure directly affects sensitivity to noise.

\begin{table}[t]
  \caption{\textbf{Linear feature recovery errors at different noise levels.} Lower error indicates better feature recovery.}
  \label{tab:recovery-error}
  \centering
  \begin{tabular}{lccccc}
    \toprule
    & \multicolumn{5}{c}{Noise level $\tau$} \\
    \cmidrule(lr){2-6}
    Representation & 0.1 & 0.3 & 0.5 & 0.7 & 0.9 \\
    \midrule
    V-RAE & 0.006 & 0.057 & 0.206 & 0.514 & 0.919 \\
    World-R0 & 0.012 & 0.129 & 0.364 & 0.655 & 0.933 \\
    \bottomrule
  \end{tabular}
\end{table}

These results suggest that, under the same noise schedule, recoverable information is lost earlier in representations whose energy is spread across many principal directions with moderate or low energy. This observation motivates the use of a gentler noise schedule to prevent feature signals from becoming noise-dominated too early.

\subsection{Feature-Energy-Based Shift Correction}

Following SD3~\cite{sd3}, RAE~\cite{rae}, and related work, DiTs that adopt the shift parameter typically determine its value based on the number of latent tokens $N$ and channels $C$

\begin{equation}
s_0
=
\sqrt{\frac{NC}{4096}}.
\end{equation}

Here $N$ denotes the number of tokens in one latent, and $C$ denotes the number of channels per token.

The time variable $t\in[0,1]$ is mapped by the shift to the actual noise intensity:

\begin{equation}
\tau(t;s)
=
\frac{st}{1+(s-1)t}.
\end{equation}

Here $s$ is the shift being used, and $\tau(t;s)$ is the coefficient applied during forward noise injection.

Along the $i$-th principal direction, the signal-to-noise ratio can be written as:

\begin{equation}
\operatorname{SNR}_i(t;s)
=
\lambda_i
\frac{(1-t)^2}{s^2t^2},
\end{equation}

where $\lambda_i$ is the feature variance along the $i$-th principal direction. The shift therefore changes signal retention along every direction at once. Our spectral and recovery analyses show that representations with similar latent sizes can still distribute energy very differently. We first correct the default shift according to the number of directions that carry effective energy.

Let the proportion of effective directions be:

\begin{equation}
\rho
=
\frac{1}{C}
\sum_{i=1}^{C}
\mathbf{1}
\left[
\lambda_i
\geq
10^{-3}\lambda_{\max}
\right],
\end{equation}

where $\lambda_{\max}=\max_i\lambda_i$ and $\mathbf{1}[\cdot]$ is an indicator function that is 1 when the condition holds and 0 otherwise. The effective number of directions is then

\begin{equation}
C_{\mathrm{eff}}
=
C\rho.
\end{equation}

$C_{\mathrm{eff}}$ is an equivalent channel count inferred from the feature-energy distribution, rather than the actual number of channels. It represents a threshold-based effective spectral dimension, counting principal directions whose eigenvalues exceed a specified threshold relative to the largest eigenvalue.

Replacing the channel count $C$ in the default shift with the effective channel count $C_{\mathrm{eff}}$ gives a shift that considers only the number of effective directions:

\begin{equation}
s_{\mathrm{eff}}
=
\sqrt{
\frac{N C_{\mathrm{eff}}}{4096}
}
=
\sqrt{
\frac{NC\rho}{4096}
}
=
s_0\sqrt{\rho}.
\end{equation}

This expression keeps the original square-root scaling with latent dimensions while adjusting the noise level to the number of directions carrying effective energy. Fewer effective directions mean a smaller $\rho$ and therefore a smaller shift.

However, simply counting the number of effective directions cannot distinguish differences in energy distribution across these directions. For example, two representations may have the same number of effective directions, while their energy is concentrated in a few directions or distributed across multiple directions. Under similar total energy and the same noise level, high-energy principal directions can maintain a higher signal-to-noise ratio (SNR), whereas moderate- and low-energy principal directions are more easily dominated by noise. Therefore, different energy distributions may lead to different levels of signal preservation and linear recovery errors~\cite{kay1993fundamentals}. To capture energy distribution differences that are not reflected by the number of effective directions, we further calculate the log-spectral width. Since eigenvalues near zero can dominate a logarithmic statistic, only values above a threshold are included:

\begin{equation}
\mathcal{I}_{\sigma}
=
\left\{
i\in\{1,\ldots,C\}
\mid
\lambda_i>10^{-6}
\right\},
\qquad
C_{\sigma}=|\mathcal{I}_{\sigma}|.
\end{equation}

\begin{equation}
\mu_{\log\lambda}
=
\frac{1}{C_{\sigma}}
\sum_{i\in\mathcal{I}_{\sigma}}
\log_{10}\lambda_i,
\end{equation}

\begin{equation}
\sigma_{\log\lambda}
=
\sqrt{
\frac{1}{C_{\sigma}}
\sum_{i\in\mathcal{I}_{\sigma}}
\left(
\log_{10}\lambda_i-\mu_{\log\lambda}
\right)^2
}.
\end{equation}

Here, $\mathcal{I}_{\sigma}$ is the index set of eigenvalues included in the spectral-width calculation, and $C_{\sigma}$ is the number of retained eigenvalues. The quantities $\mu_{\log\lambda}$ and $\sigma_{\log\lambda}$ are the mean and standard deviation of their logarithms, respectively. The latter measures the energy span of the spectrum: larger differences between principal directions produce a wider span and a larger value. Whereas $\rho$ counts how many directions retain effective energy, $\sigma_{\log\lambda}$ measures how unevenly that energy is distributed.

We therefore use the log-spectral width to further modulate the shift derived from the effective channel count:

\begin{equation}
s_{\mathrm{mod}}
=
s_{\mathrm{eff}}\sigma_{\log\lambda}
=
s_0\sigma_{\log\lambda}\sqrt{\rho}.
\end{equation}

The modulation has two parts. The factor $\sqrt{\rho}$ replaces the nominal channel count with an effective count, while $\sigma_{\log\lambda}$ captures the energy spread across principal directions. Together, $\rho$ and $\sigma_{\log\lambda}$ describe how many directions are effective and how unevenly their energy is distributed, yielding a shift specific to the representation.

When $\sigma_{\log\lambda}\sqrt{\rho}<1$, the corrected shift is smaller than the default, and the same time $t$ corresponds to weaker noise. When $\sigma_{\log\lambda}\sqrt{\rho}\approx1$, the corrected shift is close to the default configuration.

This correction is empirical: it is motivated by observations in the spectral-statistics and noise-recovery experiments and is intended to improve the match between the shift and different representations; it is not claimed to be theoretically unique or optimal. It leaves the DiT architecture and training objective unchanged and only modifies the mapping from time $t$ to the actual noise level $\tau$ during training.

\subsection{Application to Video Generation}

For video generation, we first encode the input videos with World-R0 to obtain the latents used by the generative model. We then compute the channel covariance of these latents on the training set and calculate $\sigma_{\log\lambda}$, $\rho$, and the corresponding $s_{\mathrm{mod}}$. DiT is still trained with rectified flow, with the representation-dependent shift $s_{\mathrm{mod}}$ applied only during training to adjust the noise schedule:

\begin{equation}
z_\tau=(1-\tau)z_0+\tau\varepsilon,
\qquad
\tau=\frac{s_{\mathrm{mod}}t}
{1+(s_{\mathrm{mod}}-1)t}.
\end{equation}

After training, the video decoder converts DiT's generated latents back to RGB videos. Feature statistics are used only to set the noise schedule, so neither the World-R0 encoder nor the main DiT architecture is changed. Dataset-specific statistics and results are reported in the experiments section.

\section{Extensions to World-Modeling Tasks}
\label{sec:world-model-tasks}

\subsection{Action Generation}

\begin{figure}[t]
  \centering
  \includegraphics[width=0.8\textwidth]{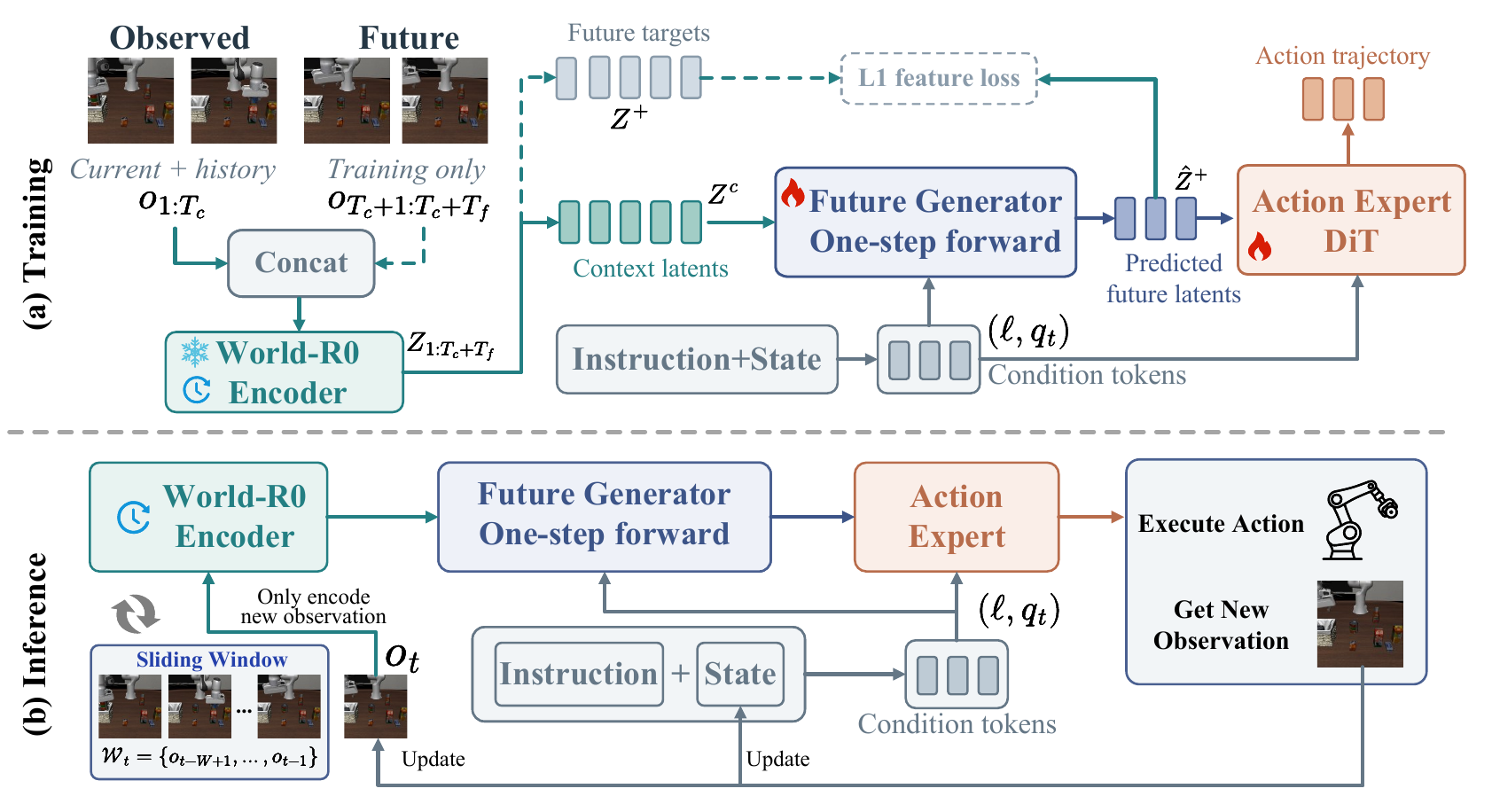}
  \caption{\textbf{World-R0-driven future visual-state prediction and robot-action generation framework.} The model extracts causal latents from current and historical observations, predicts future visual states with a Generator, and generates action trajectories with an action expert conditioned on the task and predicted states.}
  \label{fig:action}
\end{figure}

We test whether World-R0 can support robot control directly in representation space. From the current and historical observations, the model first extracts a contextual world representation. A Generator predicts the future state in one step, and an action expert decodes that predicted representation into a robot-arm trajectory. The model predicts future states in latent space and generates actions from these predictions, without first rendering future video frames and inferring scene states from them. The detailed architecture is shown in Figure~\ref{fig:action}.

Let $o_{1:T_c}$ denote the historical and current observations and $o_{T_c+1:T_c+T_f}$ the future observations, where $T_c$ and $T_f$ are the context and future-target lengths. During training, we concatenate both parts and feed the resulting sequence to the encoder:

\begin{equation}
Z_{1:T_c+T_f}
=
E_{\mathrm{causal}}
\left(
 o_{1:T_c+T_f}
\right).
\end{equation}

Because the encoder uses causal attention along time, the position at time $t$ can access only current and historical observations, and future positions do not affect the outputs at context positions. The encoded result can therefore be split into contextual and future-target latents:

\begin{equation}
Z^{c}=Z_{1:T_c},
\qquad
Z^{+}=Z_{T_c+1:T_c+T_f}.
\end{equation}

The model can access $Z^{c}$, whereas $Z^{+}$ is encoded from future observations and serves only as the training target. The task condition is $(\ell,q_t)$, with $\ell$ the language instruction and $q_t$ the current robot state.
The Generator predicts the future representation in one step:

\begin{equation}
\hat Z^{+}
=
P_{\phi}
\left(
 Z^{c}\mid \ell,q_t
\right),
\end{equation}

where $P_{\phi}$ is a Generator with parameters $\phi$ and $\hat Z^{+}$ is the predicted future latent. Future representation prediction uses the L1 loss:

\begin{equation}
\mathcal{L}_{\mathrm{future}}
=
\mathbb{E}
\left[
\left\|
\hat Z^{+}
-
\operatorname{sg}(Z^{+})
\right\|_{1}
\right].
\end{equation}

The L1 norm is taken over all future time positions, spatial positions, and channels, and $\operatorname{sg}(\cdot)$ denotes stop-gradient. Because the World-R0 encoder is frozen in this stage, $Z^{+}$ receives no gradient. The loss directly matches the predicted and true future latents, training the Generator to predict future states in representation space.

The action-generation model is a flow-matching DiT. During training, a real action segment $a_t$ and random Gaussian noise $\epsilon\sim\mathcal{N}(0,I)$ are linearly interpolated:

\begin{equation}
a_t^{\tau_a} = (1-\tau_a)\epsilon+\tau_a a_t, \qquad \tau_a\in[0,1].
\end{equation}

The parameter $\tau_a$ controls the noise level. Given the noisy action $a_t^{\tau_a}$, time $\tau_a$, task condition $(\ell,q_t)$, and Generator output $\hat Z^{+}$, the action expert predicts the target velocity $a_t-\epsilon$:

\begin{equation}
\hat v_t = v_{\psi} \left( a_t^{\tau_a},\tau_a; \hat Z^{+},\ell,q_t \right).
\end{equation}

The corresponding flow-matching loss is:

\begin{equation}
\mathcal{L}_{\mathrm{action}} = \mathbb{E} \left[ \left\| \hat v_t-(a_t-\epsilon) \right\|_2^2 \right].
\end{equation}

Here $v_{\psi}$ denotes the action DiT with parameters $\psi$.
The final training objective is:

\begin{equation}
\mathcal{L}
=
\mathcal{L}_{\mathrm{action}}
+
\lambda_{\mathrm{future}}
\mathcal{L}_{\mathrm{future}},
\end{equation}

where $\lambda_{\mathrm{future}}$ controls the weight of future-representation prediction supervision.

During inference, to fully exploit the advantages of causal encoding, the model maintains a fixed-length temporal window rather than accumulating historical observations indefinitely. Let the window size be $W$; the context at the current step consists of the most recent $W$ temporal units:

\begin{equation}
\mathcal{W}_t
=
\left\{
 o_{t-W+1},\ldots,o_t
\right\}.
\end{equation}

When a new observation arrives, the encoder reuses computation for existing positions in the window through the KV cache and applies causal encoding only to the new observation. Historical information can therefore be continuously reused within a finite context.

At each decision step, the Generator predicts $\hat Z^{+}$ from the current window. The action expert starts from Gaussian noise and integrates the flow-matching trajectory to produce a future action segment. Once the robot executes part of that segment and receives a new observation, the model updates the window and KV cache and predicts again. The resulting loop alternates causal encoding, future-state prediction, action generation, and window updates.

The implementation is close to V-JEPA-Policy~\cite{vjepa-policy}: both predict future visual features and then generate actions from those predictions. V-JEPA-Policy, however, uses a bidirectional encoder to form a single context latent. A new observation can change the representations of earlier observations, which makes incremental sliding-window updates with a KV cache difficult. World-R0's causal encoder keeps each temporal representation independent of future observations, so the cache can be updated within a fixed-length window by processing only the newly arrived data. On LIBERO, this design yields better control performance than V-JEPA-Policy; detailed results are in Section~\ref{sec:exp-action-generation}.

\subsection{3D Gaussian Scene Generation}

\begin{figure}[t]
  \centering
  \includegraphics[width=\textwidth]{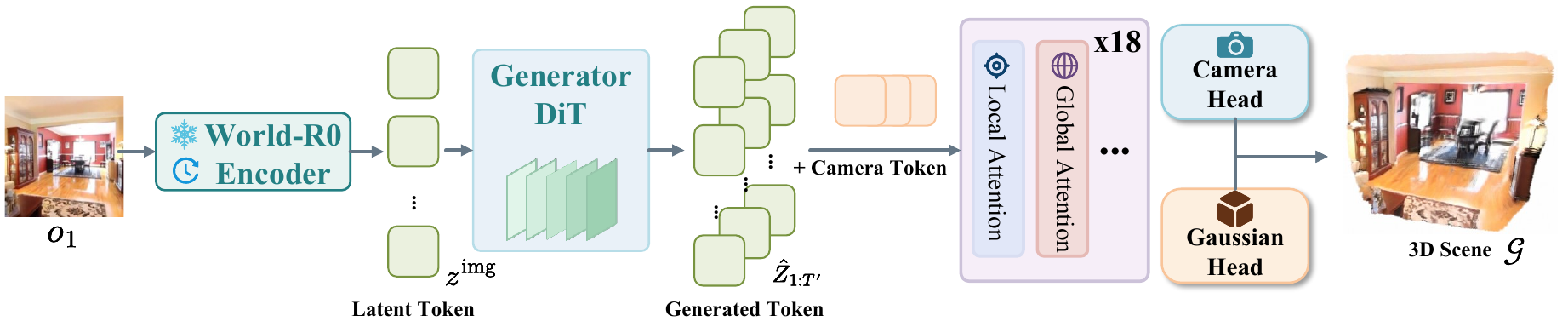}
  \caption{\textbf{World-R0 single-image-conditioned 3D Gaussian scene generation pipeline.} The input image is duplicated and passed through the causal encoder to obtain the conditioning latent. A video DiT generates the complete latent sequence from this condition, and a Gaussian decoder converts it into a renderable 3D Gaussian scene.}
  \label{fig:3dgs}
\end{figure}

We also use World-R0 for single-image-conditioned 3D Gaussian scene generation~(Figure~\ref{fig:3dgs}). The goal is to decode World-R0 latents directly into 3DGS, preserving spatial, geometric, and appearance information. The model builds a context representation from the input image, takes the latent at the first temporal position as the image condition, predicts the full latent sequence with a video DiT, and decodes that sequence into a renderable 3D Gaussian scene.

Given an input image $o_1$, we copy it at the beginning of a video sequence and construct a sequence with a repeated first frame:

\begin{equation}
\tilde o_{1:T}
=
[o_1,o_1,o_2,\ldots,o_{T-1}].
\end{equation}

After temporal compression and causal encoding, we obtain the compressed video latent $Z_{1:T'}$.

Because the first temporal unit depends only on the repeated image at the beginning of the sequence, the first latent can serve as the single-image condition:

\begin{equation}
z^{\mathrm{img}}=Z_1.
\end{equation}

The latent $z^{\mathrm{img}}$ is determined only by the input image, whereas $Z_{1:T'}$ represents the complete compressed video. Replicating the first frame and using causal encoding ensure that the image condition contains no information from later frames.

The training process has two stages. The first stage trains the Gaussian decoder. We use the pose-free reconstruction framework of YoNoSplat~\cite{yonosplat}, whose decoder consists of alternating-attention modules, a Gaussian head, and a camera head. With the World-R0 encoder frozen, the complete video is encoded into $Z_{1:T'}$, and the Gaussian decoder outputs a 3D scene:

\begin{equation}
\mathcal{G}
=
D_{\mathrm{G}}
\left(
 Z_{1:T'}
\right).
\end{equation}

$D_{\mathrm{G}}$ denotes the Gaussian decoder, and $\mathcal{G}$ is a scene representation made up of 3D Gaussian centers, scales, orientations, opacities, and colors or features. This stage teaches the decoder to read spatial structure and appearance from World-R0 latents.

The second stage trains an image-conditioned video DiT. Given the first-frame condition $z^{\mathrm{img}}$, DiT predicts the complete video latent:

\begin{equation}
\hat Z_{1:T'}
=
G_{\theta}
\left(
z^{\mathrm{img}}
\right).
\end{equation}

$G_{\theta}$ is the video DiT with parameters $\theta$, and $\hat Z_{1:T'}$ is the complete sequence generated from the image condition. During training, the target latents $Z_{1:T'}$ are extracted from real videos by the World-R0 encoder. DiT learns to predict latents at subsequent time steps conditioned on the first-frame representation. The Gaussian decoder is frozen in this stage.

At inference, the model receives a single image. World-R0 first extracts $z^{\mathrm{img}}$; the video DiT then generates $\hat Z_{1:T'}$, and the Gaussian decoder converts that sequence into a 3D scene.
The pipeline does not require an intermediate RGB video, multiple input frames, or explicit camera-pose inputs. On RE10K it achieves a gFVD of 93.21, demonstrating that World-R0 latents can effectively support 3D scene generation from a single image.

\section{Experiments and Analysis}

\subsection{Representation Adaptation and Performance Analysis}

\subsubsection{Experimental Setup}

\textbf{Data and input settings.} For the representation comparison and related ablations, we use about 150,000 training samples, 688$\times$368 inputs, and 49-frame continuous clips with first-frame padding. For the data-scaling experiment, we increase the training set to about 2.4 million samples, switch to 256$\times$256 inputs, and use 16 frames for the final model. The 49-frame setup is used only to choose the adaptation method and run the ablations. After each dataset is split into clips and repeatedly sampled, the resulting numbers of training entries are listed in Table~\ref{tab:data-comp}.

\begin{table}[t]
  \caption{\textbf{Data composition for the two representation-learning configurations.} ``--'' indicates that the dataset is not used in that configuration.}
  \label{tab:data-comp}
  \centering
  \begin{tabular}{lcc}
    \toprule
    Dataset & Ablation config. & Full config. \\
    \midrule
    ACID~\cite{acid} & -- & 6,224 \\
    DL3DV~\cite{dl3dv} & 20,931 & 20,931 \\
    Pexels-45K~\cite{pixels} & -- & 29,135 \\
    Kinetics-400 (K400)~\cite{kinetics} & 20,000 & 240,436 \\
    RealEstate10K (RE10K)~\cite{re10k} & 58,515 & 58,515 \\
    SpatialVid~\cite{spatialvid} & -- & 363,668 \\
    ScanNet++~\cite{scannetpp} & 9,817 & 39,268 \\
    Something-Something V2 (SSv2)~\cite{ssv2} & 40,000 & 168,913 \\
    UCF101~\cite{ucf101} & -- & 9,537 \\
    PVD~\cite{PVD} & -- & 998,812 \\
    Kinetics-600 (K600)~\cite{k600} & -- & 426,253 \\
    YouTube-VOS~\cite{youtubevos} & 3,471 & 17,355 \\
    \midrule
    \textbf{Total} & \textbf{152,734} & \textbf{2,379,047} \\
    \bottomrule
  \end{tabular}
\end{table}

\textbf{Training hyperparameters.} The three adaptation methods use the following key hyperparameters. Each temporal-autoencoding loss has weight $1.0$. For feature prediction, the masked-block aspect ratio is sampled from $[0.75,1.5]$, with 8 blocks at scale $0.15$ and 2 blocks at scale $0.7$; the same spatial region is masked across consecutive time steps. The masked-prediction, context-consistency, and Gram losses have weights $1.0$, $0.5$, and $0.2$. Asymmetric distillation uses the same masks. The video and image teacher alignment losses have weights $1.0$ and $0.5$, while masked and visible positions have weights $1.0$ and $0.5$, yielding $\alpha_{\mathrm{v}}=1.0$, $\beta_{\mathrm{v}}=0.5$, $\alpha_{\mathrm{i}}=0.5$, and $\beta_{\mathrm{i}}=0.25$.

\textbf{Frozen probing setup.} In every probing experiment, the visual encoder is frozen; only the task head and any necessary temporal upsampling module are trained. World-R0 uses the final-layer features of the EMA causal ViT-L, with 1,024 channels and a spatial patch size of 16. We apply ImageNet normalization and do not use a JEPA mask during encoding. With 2$\times$ and 4$\times$ temporal compression, 16 input frames produce 8 and 4 latents. Classification heads use global mean pooling over the spatiotemporal features, whereas dense heads upsample temporally to the input frame rate and predict frame by frame. Table~\ref{tab:probe-setup} gives the input and optimization settings for the two training configurations. The ablation configuration uses 150,000 samples at 688$\times$368, and the full configuration uses 2.4 million samples at 256$\times$256. Task definitions and metrics are unchanged; only the input dimensions vary with resolution. Table~\ref{tab:probe-setup} lists the specific input configuration for each task. 

\begin{table}[t]
  \caption{\textbf{Training settings for frozen probing tasks.} The ``ablation configuration'' uses 150,000 samples and 688$\times$368 resolution; the ``full configuration'' uses 2.4 million samples and 256$\times$256 resolution.}
  \label{tab:probe-setup}
  \centering
  \resizebox{\textwidth}{!}{
  \setlength{\tabcolsep}{3pt}
  \begin{tabular}{lccccccccc}
    \toprule
    Task & \shortstack{Ablation\\config.} & \shortstack{Full\\config.} & Optimizer & Init. LR & Weight decay & Training budget & Warmup & Batch size & Clip \\
    \midrule
    K400 & \shortstack{16 frames,\\224$\times$224 crop} & \shortstack{16 frames,\\256$\times$256 crop} & SGD & 0.1 & 0 & 100 epochs & 10 epochs & 2048 & 1.0 \\
    SSv2 & \shortstack{16 frames,\\224$\times$224 crop} & \shortstack{16 frames,\\256$\times$256 crop} & SGD & 0.01 & 1e-4 & 200 epochs & 20 epochs & 512 & 1.0 \\
    YouTube-VOS & \shortstack{16-frame clip,\\688$\times$368} & \shortstack{16-frame clip,\\256$\times$256} & AdamW & 1e-3 & 1e-4 & 20 epochs & 2 epochs & 8 & 1.0 \\
    ScanNet++ & \shortstack{16 consecutive\\frames, 688$\times$368} & \shortstack{16 consecutive\\frames, 256$\times$256} & AdamW & 1e-3 & 1e-4 & 30 epochs & 2 epochs & 128 & 1.0 \\
    RE10K & \shortstack{Full video, 16-frame\\clips, 688$\times$368} & \shortstack{Full video, 16-frame\\clips, 256$\times$256} & AdamW & 1e-3 & 1e-4 & 30,000 steps & 3,000 steps & 16 & 5.0 \\
    \bottomrule
  \end{tabular}
  }
\end{table}

\subsubsection{Main Results}

\begin{table}[t]
  \caption{\textbf{Comparison of visual representations and World-R0.} IDs correspond to the analysis in the text. Both World-R0 versions use the final data configuration. Bold and underline indicate the best and second-best results, respectively, within the video representation group (\#5–\#8).}
  \label{tab:repr-compare}
  \centering
  \resizebox{\textwidth}{!}{
  \begin{tabular}{lllccccccc}
    \toprule
    ID & Method & Backbone & Comp. ratio & Causal &
    \begin{tabular}[t]{@{}c@{}}K400\\Top-1 acc. $\uparrow$\end{tabular} &
    \begin{tabular}[t]{@{}c@{}}SSv2\\Top-1 acc. $\uparrow$\end{tabular} &
    \begin{tabular}[t]{@{}c@{}}YouTube-VOS\\J\&F $\uparrow$\end{tabular} &
    \begin{tabular}[t]{@{}c@{}}ScanNet++\\AbsRel $\downarrow$\end{tabular} &
    \begin{tabular}[t]{@{}c@{}}RE10K\\PSNR $\uparrow$\end{tabular} \\
    \midrule
    \#1 & RAEv2 & DINOv3-L & 1 & -- & 76.30 & 29.64 & 0.6016 & 0.2082 & 24.19 \\
    \rowcolor{gray!13}
    \#2 & DINOv3-L & -- & 1 & -- & 76.51 & 29.92 & 0.5888 & 0.2051 & 21.56 \\
    \midrule
    \#3 & FAE & DINOv2-G & 1 & -- & 59.12 & 14.48 & 0.6813 & 0.2492 & 20.02 \\
    \rowcolor{gray!13}
    \#4 & DINOv2-G & -- & 1 & -- & 77.60 & 30.09 & 0.5589 & 0.2161 & 21.01 \\
    \midrule
    \#5 & V-RAE & V-JEPA2.1-L & 4 & $\times$ & 59.25 & 37.46 & 0.4950 & 0.2411 & 24.60 \\
    \rowcolor{gray!13}
    \#6 & V-JEPA2.1-L & -- & 2 & $\times$ & \underline{69.96} & \underline{51.44} & \textbf{0.6111} & \textbf{0.2008} & 24.32 \\
    \#7 & World-R0 & V-JEPA2.1-L & 2 & \checkmark & \textbf{70.03} & \textbf{52.59} & \underline{0.5791} & \underline{0.2160} & \textbf{25.78} \\
    \#8 & World-R0 & V-JEPA2.1-L   & 4 & \checkmark & 67.69 & 51.13 & 0.5146 & 0.2202 & \underline{25.47} \\
    \bottomrule
  \end{tabular}
  }
\end{table}

Table~\ref{tab:repr-compare} presents the results of different foundation representations and adapted representations on five frozen probing tasks.
We first compare the differences among the foundation representations. DINOv3-L (\#2) and DINOv2-G (\#4) are both primarily designed for single-image representation learning and therefore exhibit strong performance in static semantic understanding. In contrast, the bidirectional V-JEPA2.1-L (\#6) significantly outperforms both image representations on SSv2, while also achieving better results in depth estimation, video segmentation, and appearance reconstruction. This is related to the dense prediction strategy of V-JEPA2.1: during training, supervision is applied to both visible and masked spatiotemporal tokens, with prediction constraints imposed on multi-layer features. This enables the model to learn temporal dynamics while preserving local spatial structure and appearance information.

To examine how adaptation affects the information retained in foundation representations, we compare RAEv2 and FAE with their respective backbones, in addition to the V-RAE comparison discussed above. RAEv2 (\#1) directly aggregates the features from seven layers of DINOv3-L and trains a decoder for image reconstruction, while keeping the visual encoder frozen throughout training. Compared with the original DINOv3-L representation (\#2), \#1 achieves comparable results in classification, video segmentation, and depth estimation, with improved appearance reconstruction, largely preserving multiple capabilities of the original representation. Compared with the original DINOv2-G representation (\#4), FAE (\#3) improves only on YouTube-VOS, while showing performance degradation on the other four probing tasks. This indicates that feature compression and reconstruction lead to information loss across multiple aspects. The improvement on YouTube-VOS may be attributed to FAE's bottleneck structure (the channel-compression module used in the AE encoder) filtering out high-frequency details, making object-level structures more prominent in the representation.

V-RAE (\#5) adapts V-JEPA2.1-L, further reducing the number of temporal tokens and requiring the representation to preserve the original video information within a more compact latent space. Compared with the bidirectional V-JEPA2.1-L (\#6), V-RAE shows performance degradation on multiple tasks, indicating that both temporal compression and the adaptation process introduce some information loss. In essence, adapting an existing video representation involves reorganizing the original features and compressing their representation space, making it difficult to fully preserve all semantic, temporal, spatial, and appearance information contained in the original backbone.

The post-training of World-R0 uses teacher representation supervision to adapt an existing video foundation representation into a compact latent that supports causal updates. Under the 2$\times$ compression setting, \#7 achieves 70.03\% on K400 and 52.59\% on SSv2, improving over the bidirectional V-JEPA2.1-L (\#6) by 0.07 and 1.15 percentage points, respectively. The PSNR on RE10K also increases from 24.32 dB to 25.78 dB. These improvements are mainly attributed to the complementary priors provided by larger and more capable video and image teachers. The former contains cross-frame dynamic information, while the latter supplements frame-level semantics and appearance information. Together, they enable the causal student to retain part of the teachers' representational capabilities even when it can only access current and past frames. On the other hand, the J\&F score on YouTube-VOS decreases from 0.6111 to 0.5791, and the AbsRel on ScanNet++ increases from 0.2008 to 0.2160. Since causal encoding cannot access future frames, some performance degradation remains in segmentation and depth estimation tasks that rely on complete spatiotemporal context. Overall, World-R0 after causal adaptation maintains comparable performance to the original bidirectional representation across the five tasks, demonstrating that teacher supervision effectively mitigates representation degradation caused by causal constraints.

Under the 4$\times$ compression setting, \#8 achieves improvements across all five probing tasks compared with V-RAE (\#5) under the same temporal compression ratio. This indicates that, compared with directly adding adaptation modules to an existing video representation, post-training continuously constrains the student representation during compression and causal adaptation, reducing information loss caused by shifts in the feature distribution. In summary, World-R0 maintains a favorable balance across semantic, temporal, spatial, and appearance tasks even at high compression ratios.

\subsubsection{Ablation Experiments}
\label{sec:ablation}

Table~\ref{tab:adapt-abl} includes ablations of the three representation adaptation methods and teacher representations. \#1–\#6 evaluate different adaptation methods. \#1 is the reproduced V-JEPA2.1-L-causal baseline. \#2 and \#3 use temporal feature autoencoding and feature prediction adaptation, respectively. \#4 and \#5 compare different weights for two-teacher distillation, and \#6 uses a single video teacher. \#7–\#9 directly evaluate the teacher representations and their combination to analyze complementary information from the image and video teachers. All adaptation ablations use about 150,000 samples, 688$\times$368 input resolution, and 49-frame continuous sequences. Classification probing uses 224$\times$224 inputs, and dense probing uses 688$\times$368 inputs. Because probing resolution and training scale differ, the metrics in this table cannot be directly compared with those in Table~\ref{tab:repr-compare}.

\textbf{Ablation of representation adaptation methods.} In \#1, we replace the bidirectional temporal attention in the original model with causal attention, initialize it with the weights of the original bidirectional model, and train it according to the official V-JEPA code and procedure. This baseline shows that directly replacing bidirectional attention with causal attention causes a clear loss in representational capabilities.

For \#2, the reconstruction targets are the input DINO features and their local temporal changes. This directly preserves current-frame content and allows the method to inherit DINO's frame-level semantics, yielding 76.45\% on K400. However, the temporal-change loss only constrains local differences between adjacent frames; it does not require recovery of unseen spatiotemporal units from context. Long-range dependencies, action changes, and object correspondence are therefore weakly modeled, as reflected by SSv2 at 26.11\% and YouTube-VOS J\&F at 0.4367. These results suggest that direct reconstruction is better suited to preserving single-frame semantics but is insufficient to capture the spatiotemporal structure of continuous observations.

For \#3, the target is a masked latent recovered from visible spatiotemporal context. The model must exploit relations across both time and space, which strengthens cross-frame dependencies and spatiotemporal modeling. Relative to \#2, SSv2 rises by 13.18 points, YouTube-VOS J\&F by 0.1711, and RE10K PSNR by 0.86 dB, indicating better modeling of action changes and global appearance structure. The target, however, emphasizes information predictable from context and imposes a weaker direct constraint on frame-level discriminative features. K400 therefore falls from 76.45\% to 72.14\%, while ScanNet++ AbsRel rises from 0.2135 to 0.2202. This method shifts the representation toward spatiotemporal relations at the cost of some frame-level semantic separability and fine-grained geometry.

The asymmetric distillation model in \#4 improves several capabilities, especially cross-frame action understanding, video segmentation, geometric estimation, and appearance reconstruction. Its K400 score is lower than those of \#2 and \#3, suggesting that the student prioritizes broad spatiotemporal information over single-frame classification. Because \#4 learns mainly by predicting teacher features from visible context rather than directly reconstructing frame-level features, it receives weaker direct supervision for frame-level information, which may partly explain its lower K400 classification accuracy. The two teachers complement one another, allowing \#4 to retain more temporal, spatial, and appearance information under causal access and to achieve a more balanced profile across tasks.

\textbf{Analysis of teacher-representation combinations.} Results for \#7 and \#8 show that DINOv3-H favors static semantics and appearance, whereas V-JEPA2.1-g favors cross-frame relations and spatial structure. Their combination in \#9 gives the lowest AbsRel on ScanNet++, but falls between the two single-teacher results on K400, SSv2, YouTube-VOS, and RE10K rather than matching the best of either. Combining the two teacher representations therefore provides complementary information and yields a more balanced performance profile across probing tasks. These results provide a reference for assessing the complementary properties of the teacher representations.

\textbf{Ablation of distillation-teacher sources.} Using only the video teacher, \#6 achieves 49.16\% on SSv2 and 0.1869 on ScanNet++, surpassing \#4's 46.30\% and 0.1979. Its K400 score, 63.73\%, is almost identical to \#4's 63.85\%. On YouTube-VOS and RE10K, however, \#6 reaches J\&F 0.6522 and PSNR 27.41 dB, below \#4's 0.6607 and 28.01 dB. The video teacher is therefore especially helpful for cross-frame action and geometry, while the image teacher improves spatial correspondence and appearance reconstruction. We use the two-teacher setting in \#4 for the best overall balance.

\textbf{Ablation of two-teacher weights.} \#4 and \#5 share the same architecture, data, and training budget; only the teacher-loss weights differ. \#4 uses a video:image ratio of 1:0.5, while \#5 uses 1:1. Increasing the image teacher's weight improves depth estimation but degrades performance on classification, action recognition, video segmentation, and appearance reconstruction. The video teacher emphasizes dynamics and temporal relations, whereas the image teacher emphasizes static structure and appearance. The 1:0.5 ratio retains strong temporal supervision while adding complementary image information and produces the best balance across tasks.

\begin{table}[t]
  \caption{\textbf{Ablation results for representation adaptation methods.} Bold and underline indicate the best and second-best results, respectively, among the adapted representations (\#1–\#6).}
  \label{tab:adapt-abl}
  \centering
  \resizebox{\textwidth}{!}{
  \begin{tabular}{lllccccc}
    \toprule
    ID & Method & Backbone &
    \begin{tabular}[t]{@{}c@{}}K400\\Top-1 acc. $\uparrow$\end{tabular} &
    \begin{tabular}[t]{@{}c@{}}SSv2\\Top-1 acc. $\uparrow$\end{tabular} &
    \begin{tabular}[t]{@{}c@{}}YouTube-VOS\\J\&F $\uparrow$\end{tabular} &
    \begin{tabular}[t]{@{}c@{}}ScanNet++\\AbsRel $\downarrow$\end{tabular} &
    \begin{tabular}[t]{@{}c@{}}RE10K\\PSNR $\uparrow$\end{tabular} \\
    \midrule
    \#1 & Base (V-JEPA2.1-L-Causal) & V-JEPA2.1-L & 58.49 & 44.25 & 0.6498 & 0.2076 & 27.01 \\
    \#2 & Temporal feature autoencoder & DINOv3-L & \textbf{76.45} & 26.11 & 0.4367 & 0.2135 & 26.48 \\
    \#3 & Feature prediction adapter & DINOv3-L & \underline{72.14} & 39.29 & 0.6078 & 0.2202 & 27.34 \\
    \#4 & Asymmetric distillation post-training (2 teachers), weight 1:0.5 & V-JEPA2.1-L & 63.85 & \underline{46.30} & \textbf{0.6607} & 0.1979 & \textbf{28.01} \\
    \#5 & Asymmetric distillation post-training (2 teachers), weight 1:1 & V-JEPA2.1-L & 63.23 & 45.62 & \underline{0.6582} & \underline{0.1931} & \underline{27.78} \\
    \#6 & Asymmetric distillation post-training (single V-JEPA teacher) & V-JEPA2.1-L & 63.73 & \textbf{49.16} & 0.6522 & \textbf{0.1869} & 27.41 \\
    \midrule
    \rowcolor{gray!13}
    \multicolumn{8}{c}{Teacher} \\
    \#7 & DINOv3-H & \multicolumn{1}{c}{--} & 81.59 & 25.76 & 0.6654 & 0.1849 & 26.22 \\
    \#8 & V-JEPA2.1-g & \multicolumn{1}{c}{--} & 69.36 & 51.36 & 0.7130 & 0.1734 & 25.55 \\
    \#9 & V-JEPA2.1-g + DINOv3-H & \multicolumn{1}{c}{--} & 75.42 & 50.19 & 0.6901 & 0.1730 & 25.89 \\
    \bottomrule
  \end{tabular}
  }
\end{table}

\textbf{Data-scale ablation.} Under the 256$\times$256 representation-training configuration, we increase the data for asymmetric distillation post-training from about 150,000 to 600,000 and 2.4 million samples. All three models are trained for 20 epochs with the same configuration. Increasing the dataset size therefore expands sample coverage and increases the total number of training examples processed. All probing tasks use 256$\times$256 inputs, including K400 and SSv2.

\begin{figure}[t]
  \centering
  \includegraphics[width=\textwidth]{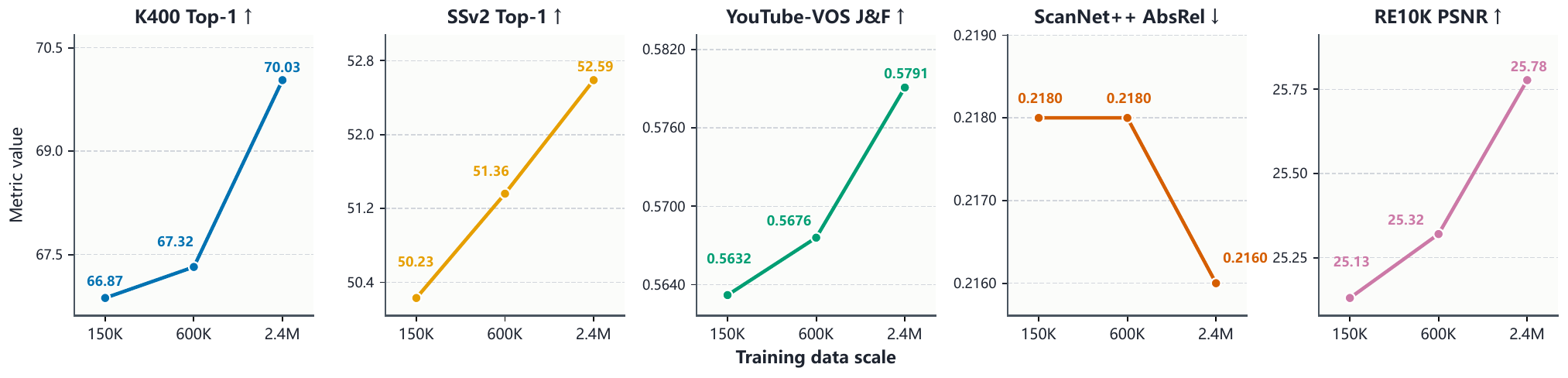}
  \caption{\textbf{Probing results of asymmetric distillation post-training at different data scales.} All three data-scale settings use 256$\times$256 resolution.}
  \label{fig:scaling}
\end{figure}
As Figure~\ref{fig:scaling} shows, all five probing metrics improve as the training set grows from 150,000 to 2.4 million samples. K400 and SSv2 rise by 3.16 and 2.36 percentage points; YouTube-VOS J\&F increases from 0.5632 to 0.5791; RE10K PSNR gains about 0.65 dB; and ScanNet++ AbsRel falls from 0.2180 to 0.2160. Classification benefits the most, while video segmentation and appearance reconstruction also improve steadily and depth estimation changes only modestly. Within the range studied here, asymmetric distillation continues to improve representational capabilities as data scale increases, with no clear sign of saturation, suggesting the potential for further scaling to more data.

\subsection{Video Generation}

\subsubsection{Experimental Setup}

\textbf{Decoder configuration.} To map World-R0 latents back to RGB videos, we adopt the V-RAE decoder and train it in two stages. The World-R0 encoder stays frozen throughout, and we fuse features from layers 11, 13, 15, 17, 19, 21, and 23. The decoder has 28 layers, 16 attention heads, hidden dimension 1,152, and tubelet size 4. Its reconstruction objective combines L1, VGG-LPIPS, Gram, temporal-difference, and I3D feature-matching losses with weights 1, 1, 100, 1, and 1.

The first stage pretrains RGB reconstruction using real-video latents from UCF101 and K600. Training uses AdamW with an initial learning rate of $5\times10^{-5}$, 3,000 warmup steps followed by cosine scheduling, 50 epochs, a global batch size of 128, bf16 precision, and EMA decay 0.9995.

In the second stage, we train a separate decoder for UCF101, K600, and Diving48 with both reconstruction and adversarial objectives. For each clean latent $z$, we sample $t\sim U[0,0.8]$ and form the forward-noised latent:

\begin{equation}
z_t=(1-t)z+t\varepsilon,
\qquad
\varepsilon\sim\mathcal{N}(0,I).
\end{equation}

We feed $z_t$ to the frozen class-conditioned DiT and denoise it to obtain $\hat z$, approximating the latent distribution encountered during generation. With probability 0.8, training mixes the clean and denoised latents:

\begin{equation}
z_{\mathrm{in}} = (1-\alpha)z+\alpha\hat z, \qquad \alpha\sim U[0,1].
\end{equation}

The corresponding real video is used as the reconstruction target, and a discriminator supplies the adversarial loss. We apply that loss only to unmixed clean latents so that distribution augmentation does not interfere with the adversarial objective.

\textbf{Generation configuration.} We use DiT$^{\mathrm{DH}}$ XL/1 for class-conditioned generation in the World-R0 latent space. The two model components have hidden dimensions of 1,536 and 2,048, depths of 28 and 2, and 24 and 16 attention heads, respectively. The encoder branch has a depth of 8. Training uses offline-extracted latents and estimates channel-normalization statistics under the same protocol. DiT uses an $x$-prediction flow-matching objective and class-conditioning dropout of 0.1. All three datasets use AdamW with $\beta=(0.9,0.95)$, weight decay 0, gradient clipping threshold 1.0, EMA decay 0.9995, bf16 precision, and random seed 3407. During training, we sample 17 frames from each video and pad the first frame three times to form a 20-frame input. At evaluation time, the model directly generates 20 frames, and the last 17 frames are used for evaluation. Training data use center crops without horizontal flips (Table~\ref{tab:dit-config}). In particular, we use 47 of the 48 Diving48 classes, excluding one class with insufficient samples; all methods are trained and evaluated on the same 47-class subset. The shift for each dataset is obtained from the feature statistics in Section~\ref{sec:feature-spectrum}.

\begin{table}[t]
  \caption{\textbf{Class-conditioned DiT training configurations for each dataset.} A linear learning rate decay schedule is used.}
  \label{tab:dit-config}
  \centering
  \begin{tabular}{lccc}
    \toprule
    Configuration & UCF101 & Diving48 & K600 \\
    \midrule
    Number of conditional classes & 101 & 47 & 600 \\
    Global batch size & 96 & 96 & 128 \\
    Training epochs & 1700 & 1500 & 100 \\
    Peak learning rate & 2e-4 & 2e-4 & 1e-4 \\
    End-of-schedule learning rate & 1e-4 & 1e-4 & 5e-5 \\
    LR decay end (epoch) & 1000 & 1000 & 100 \\
    Warmup (epoch) & 10 & 10 & 2 \\
    Training Shift & 7.746 & 8.665 & 6.918 \\
    Evaluation population & 2000 & 2115 & 6000 \\
    \bottomrule
  \end{tabular}
\end{table}

\subsubsection{Video Reconstruction Results}

Table~\ref{tab:recon} compares the reconstruction performance of two categories of video representations: Video VAEs / Tokenizers and RAE-based Methods. Within the RAE-based Methods group, World-R0 outperforms V-RAE on all four metrics on UCF101. On K600, it achieves better PSNR, SSIM, and rFVD, with only a slightly worse LPIPS. Overall, World-R0 achieves better comprehensive reconstruction performance than V-RAE. Although the two methods adopt the same decoder architecture, they differ in their training strategies. Therefore, the performance gap may be jointly influenced by representational capabilities and decoder training schemes. The asymmetric distillation employed in World-R0 is designed to preserve cross-frame, spatial, and appearance information, while DiT latent blending during decoder training can be regarded as data augmentation at the representation level, potentially improving the decoder's robustness to perturbations in the input representations.

Within the Video VAEs / Tokenizers group, methods such as Wan2.1, Wan2.2, and HunyuanVideo outperform World-R0 in visual fidelity metrics, including PSNR, SSIM, and LPIPS. This may be attributed to their larger training scales and the fact that their representation learning is primarily optimized for video reconstruction, enabling them to preserve more information relevant to pixel-level appearance recovery. On the other hand, World-R0 achieves the best rFVD across all methods on K600 and ranks only slightly below Wan2.1 on UCF101, outperforming all other methods. This indicates that its reconstructed videos exhibit a feature distribution that is relatively close to that of real videos. Therefore, although World-R0 still exhibits a gap in certain visual fidelity metrics, it remains competitive in preserving the overall feature distribution of videos.

\begin{figure}[t]
  \centering
  \includegraphics[width=\textwidth]{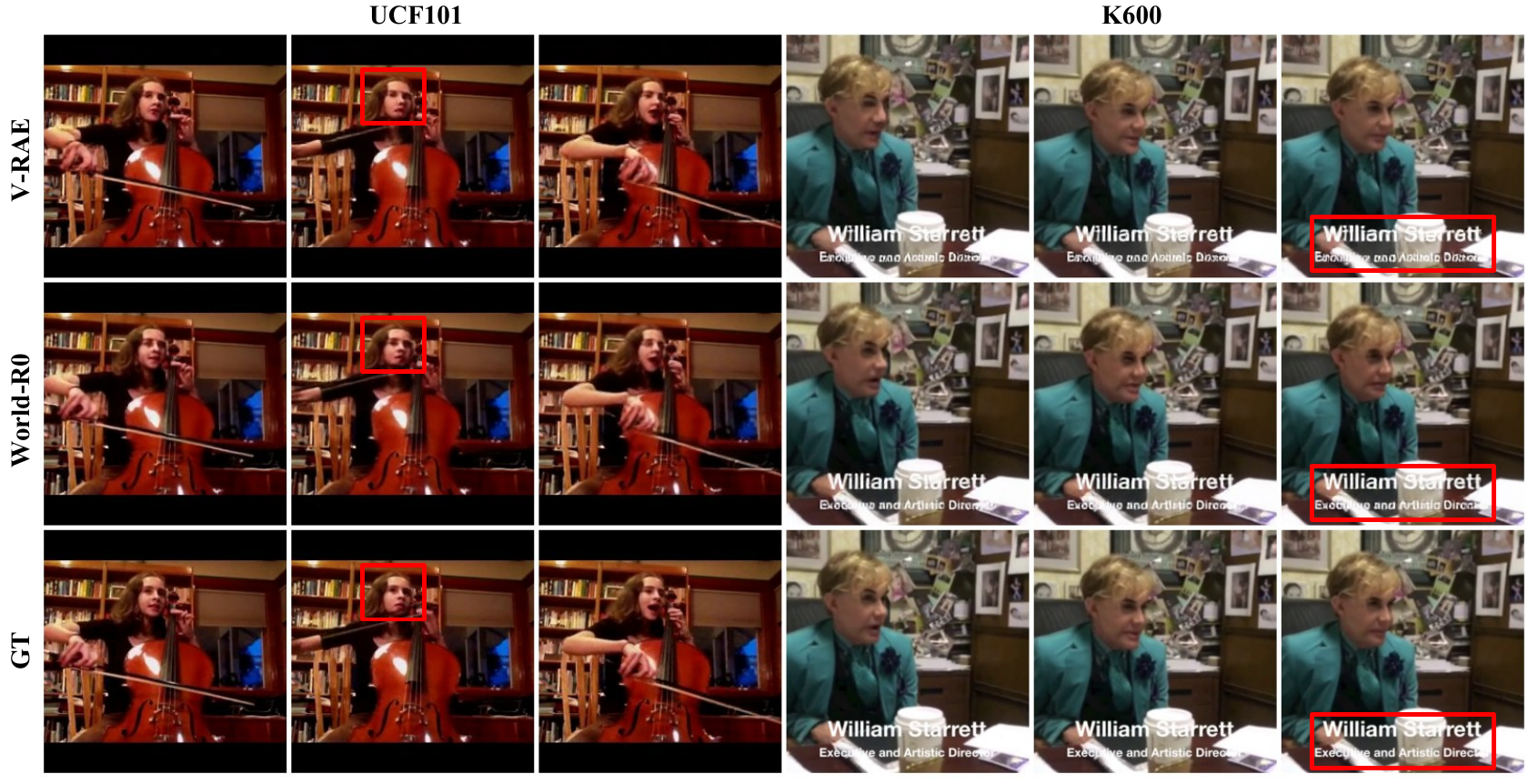}
  \caption{\textbf{Qualitative comparison of video reconstruction on UCF101 and K600.} For each dataset, the three rows show reconstructions from V-RAE and World-R0, followed by the ground-truth video frames. Red boxes highlight differences in details such as facial structures and subtitles.}
  \label{fig:recon-vis}
\end{figure}

Figure~\ref{fig:recon-vis} presents qualitative video reconstruction results on UCF101 and K600. For each dataset, the three rows correspond to the reconstruction results of V-RAE, the reconstruction results of World-R0, and the ground truth, respectively. The columns represent different frames from the same video clip, with red boxes highlighting local regions exhibiting noticeable differences. In the presented examples, World-R0 produces clearer facial structures and subtitles that are closer to the ground truth. These qualitative results are consistent with the overall quantitative trends reported in Table~\ref{tab:recon}, further demonstrating the advantages of World-R0 in recovering appearance details and local structures.

\begin{table}[t]
  \caption{\textbf{Comparison of video reconstruction results.} Lower LPIPS and rFVD are better; higher PSNR and SSIM are better. rFVD measures the Fréchet distance between the feature distributions of reconstructed and real videos. Bold and underline indicate the best and second-best results within each group, respectively.}
  \label{tab:recon}
  \centering
  \resizebox{\textwidth}{!}{
  \begin{tabular}{lccccccccccc}
    \toprule
    \multicolumn{4}{c}{\textbf{Tokenizer Configuration}}
      & \multicolumn{4}{c}{\textbf{UCF101 test}}
      & \multicolumn{4}{c}{\textbf{K600 val}} \\
    \cmidrule(lr){1-4} \cmidrule(lr){5-8} \cmidrule(lr){9-12}
    Method & Causal & Comp. & Dim.
      & LPIPS $\downarrow$ & PSNR $\uparrow$ & SSIM $\uparrow$ & rFVD $\downarrow$
      & LPIPS $\downarrow$ & PSNR $\uparrow$ & SSIM $\uparrow$ & rFVD $\downarrow$ \\
    \midrule
    \rowcolor{tabGrpA}
    \multicolumn{12}{c}{\textbf{Video VAEs / Tokenizers}} \\
    Wan2.1 VAE~\cite{wan} & \checkmark & 4 $\times$ 8 $\times$ 8 & 16 & \underline{0.068} & 34.09 & 0.937 & \textbf{6.05} & \underline{0.052} & 35.23 & 0.946 & \textbf{3.58} \\
    Wan2.2 VAE~\cite{wan} & \checkmark & 4 $\times$ 16 $\times$ 16 & 48 & \textbf{0.067} & 34.16 & 0.942 & 12.20 & \textbf{0.049} & \underline{35.73} & \underline{0.951} & 4.76 \\
    HunyuanVideo VAE~\cite{hunyuanvideo} & \checkmark & 4 $\times$ 8 $\times$ 8 & 16 & 0.069 & \textbf{35.14} & \textbf{0.950} & \underline{7.73} & 0.056 & \textbf{36.31} & \textbf{0.954} & \underline{4.38} \\
    CogVideoX VAE~\cite{cogvideox} & \checkmark & 4 $\times$ 8 $\times$ 8 & 16 & 0.081 & \underline{34.22} & \underline{0.943} & 14.53 & 0.069 & 34.82 & 0.946 & 8.11 \\
    Cosmos-0.1 (CV4$\times$8$\times$8)~\cite{cosmos} & \checkmark & 4 $\times$ 8 $\times$ 8 & 16 & 0.109 & 32.57 & 0.925 & 17.01 & 0.100 & 33.55 & 0.928 & 9.88 \\
    AToken~\cite{atoken} & $\times$ & 4 $\times$ 16 $\times$ 16 & 48 & 0.089 & 33.11 & 0.934 & 8.17 & 0.072 & 34.05 & 0.937 & 5.36 \\
    OmniTokenizer VAE~\cite{omni} & \checkmark & 4 $\times$ 8 $\times$ 8 & 8 & 0.132 & 29.59 & 0.908 & 28.86 & 0.139 & 27.75 & 0.899 & 15.35 \\
    LARP-L-long~\cite{larp} & \checkmark & 1024 tokens & 16 & 0.233 & 25.32 & 0.799 & 142.29 & 0.236 & 26.00 & 0.801 & 125.00 \\
    \cmidrule(lr){1-12}
    \rowcolor{tabGrpC}
    \multicolumn{12}{c}{\textbf{RAE-based Methods}} \\
    V-RAE (V-JEPA2.1-L)~\cite{vrae} & $\times$ & 4 $\times$ 16 $\times$ 16 & 1024 & \underline{0.121} & \underline{27.82} & \underline{0.856} & \underline{6.65} & \textbf{0.110} & \underline{28.48} & \underline{0.858} & \underline{2.13} \\
    World-R0 & \checkmark & 4 $\times$ 16 $\times$ 16 & 1024 & \textbf{0.119} & \textbf{29.56} & \textbf{0.883} & \textbf{6.18} & \underline{0.133} & \textbf{29.36} & \textbf{0.873} & \textbf{1.91} \\
    \bottomrule
  \end{tabular}
  }
\end{table}

\subsubsection{Video Generation Results}

Table~\ref{tab:ucf-gen} compares UCF101 generation across representations. Every method uses 256$\times$256 resolution, 17 output frames, and 1,280 DiT tokens, with the patch configuration adjusted to its latent space. World-R0 reaches a gFVD of 114.96, the best result among the video VAEs, AToken, and V-RAE in the table. Against V-RAE with the same DiT$^{\mathrm{DH}}$ XL/1 backbone, it reduces gFVD from 117.86 to 114.96, a reduction of 2.90. The result reflects the combined benefit of the World-R0 representation and the feature-energy-aware noise schedule.

\begin{table}[b]
  \caption{\textbf{UCF101 video-generation comparison.} ``20$\to$17'' means generation with a 20-frame configuration and evaluation over 17 frames. Bold indicates the lowest gFVD in the table.}
  \label{tab:ucf-gen}
  \centering
  \begin{tabular}{lccc}
    \toprule
    Representation & Resolution, frames & Generative model & UCF101 gFVD $\downarrow$ \\
    \midrule
    Wan2.1 VAE & 256 $\times$ 256, 17 & DiT XL/2 & 148.20 \\
    Wan2.2 VAE & 256 $\times$ 256, 17 & DiT XL/1 & 154.64 \\
    HunyuanVideo VAE & 256 $\times$ 256, 17 & DiT XL/2 & 211.53 \\
    CogVideoX VAE & 256 $\times$ 256, 17 & DiT XL/2 & 159.20 \\
    Cosmos VAE (CV4$\times$8$\times$8) & 256 $\times$ 256, 17 & DiT XL/2 & 152.70 \\
    AToken & 256 $\times$ 256, 20$\to$17 & DiT XL/1 & 143.00 \\
    \midrule
    V-RAE (V-JEPA2.1-L) & 256 $\times$ 256, 20$\to$17 & DiT$^{\mathrm{DH}}$ XL/1 & 117.86 \\
    World-R0 & 256 $\times$ 256, 20$\to$17 & DiT$^{\mathrm{DH}}$ XL/1 & \textbf{114.96} \\
    \bottomrule
  \end{tabular}
\end{table}

Table~\ref{tab:dive-k600} reports video generation results on Diving48 and K600. World-R0 reaches 66.07 on Diving48, 21.89 below V-RAE's 87.96, corresponding to a reduction of about 24.9\%. On K600 it reaches 51.51, 0.75 below V-RAE's 52.26. It therefore has the lowest gFVD among the compared methods on all three datasets.

\begin{table}[t]
  \caption{\textbf{Video-generation comparison on Diving48 and K600.} Bold indicates the lowest gFVD in each column. * indicates adaptation and reproduction using the official codebase.}
  \label{tab:dive-k600}
  \centering
  \begin{tabular}{lcc}
    \toprule
    Representation & Diving48 gFVD $\downarrow$ & K600 gFVD $\downarrow$ \\
    \midrule
    AToken* & 149.58 & 89.72 \\
    V-RAE* (V-JEPA2.1-L) & 87.96 & 52.26 \\
    World-R0 & \textbf{66.07} & \textbf{51.51} \\
    \bottomrule
  \end{tabular}
\end{table}

\begin{figure}[t]
  \centering
  \includegraphics[width=\textwidth]{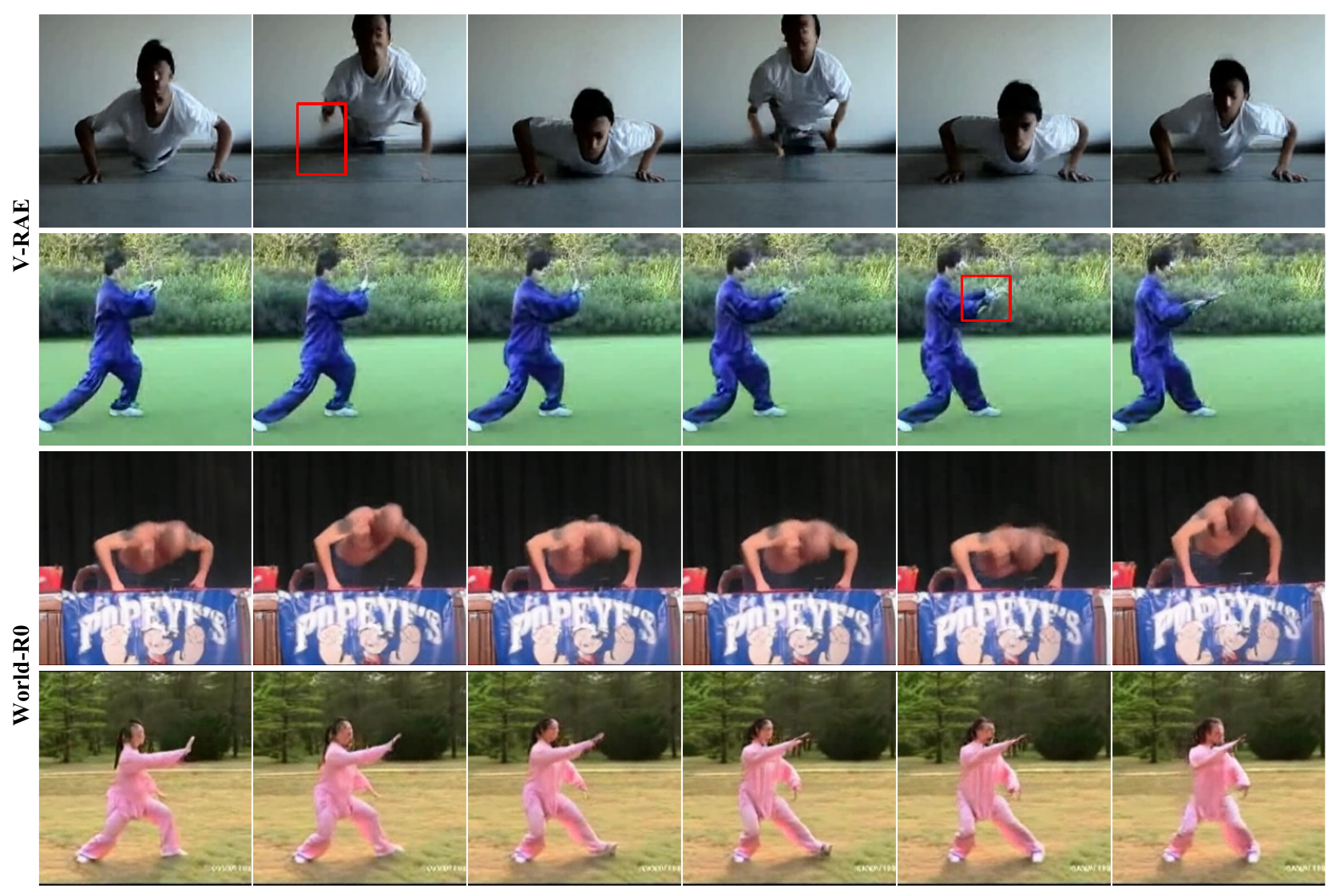}
  \caption{\textbf{Qualitative comparison of video generation by V-RAE and World-R0.} Each row shows one generated sample and columns show consecutive video frames; red boxes mark local structural artifacts.}
  \label{fig:gen-vis}
\end{figure}

Figure~\ref{fig:gen-vis} presents qualitative video-generation results for V-RAE and World-R0. Each row corresponds to one generated sample, and columns show consecutive temporal frames. V-RAE generates the correct action and overall scene but has structural instability and boundary artifacts in local motion regions such as arms and hands; red boxes mark representative errors. In contrast, World-R0 maintains more stable subject pose, limb connections, and background structure across consecutive frames, with fewer local deformations and more coherent motion.

\subsubsection{Feature-Spectrum Statistics and Shift Setting}
\label{sec:feature-spectrum}

The latents used for video generation have the same number of tokens and channels, so computing the shift only from latent dimensions gives the same default shift for the three datasets:

\begin{equation}
s_0 = \sqrt{\frac{5\cdot256\cdot1024}{4096}} \approx 17.889.
\end{equation}

We next compute a dataset-specific shift from the spectral statistics defined above. As Table~\ref{tab:spectral-stats} shows, World-R0 has a high effective-support ratio on all three datasets ($\rho=0.979$–$0.998$). Its log-spectral width ranges from 0.387 to 0.490, producing corrected shifts from 6.918 to 8.665, all well below the default. Diving48 has the broadest spectrum and the largest corrected shift; K600 has the narrowest spectrum and the smallest.

By contrast, V-RAE has effective-support ratios of only 0.358–0.396 and log-spectral widths of 1.557–1.670. Their combined corrected shifts lie between 17.216 and 17.882, close to the default value overall. Thus, even when different representations have the same latent dimensions, their eigenvalue spectra differ in log-spectral width and effective spectral support, resulting in different
corrected shift values. 

\begin{table}[t]
  \caption{\textbf{Feature-spectrum statistics and corrected shifts for World-R0 and V-RAE on the three datasets.}}
  \label{tab:spectral-stats}
  \centering
  \begin{tabular}{lcccccc}
    \toprule
    & \multicolumn{3}{c}{World-R0} & \multicolumn{3}{c}{V-RAE} \\
    \cmidrule(lr){2-4} \cmidrule(lr){5-7}
    Dataset & $\rho$ & $\sigma_{\log\lambda}$ & $s_{\mathrm{mod}}$ & $\rho$ & $\sigma_{\log\lambda}$ & $s_{\mathrm{mod}}$ \\
    \midrule
    K600 & 0.998 & 0.387 & 6.918 & 0.396 & 1.557 & 17.542 \\
    UCF101 & 0.993 & 0.435 & 7.746 & 0.369 & 1.584 & 17.216 \\
    Diving48 & 0.979 & 0.490 & 8.665 & 0.358 & 1.670 & 17.882 \\
    \bottomrule
  \end{tabular}
\end{table}

\subsubsection{Ablation of Different Shifts}

When training and inference use the same shift, s = 7.746, 17.889, and 35 yield gFVDs of 115.68, 116.25, and 125.32, respectively, indicating that smaller matched shifts generally lead to better generation quality. The best result uses 7.746 for training and 17.889 for inference, with a gFVD of 114.96. Compared with using 7.746 for both training and inference, this configuration reduces gFVD from 115.68 to 114.96; compared with using 17.889 for both stages, it reduces gFVD by 1.29.
With the training shift fixed at 7.746, increasing the inference shift to 17.889 lowers gFVD from 115.68 to 114.96, but further increasing it to 35 raises gFVD to 116.02. These results suggest that a moderate increase in the inference shift can improve generation quality, whereas further increasing it may be detrimental.

A plausible explanation is that training and inference shifts serve different purposes. World-R0 distributes its feature energy across many principal directions with moderate or low energy. These directions have relatively low signal-to-noise ratios at a given noise level and become noise-dominated earlier as the noise level increases. A smaller training shift increases the proportion of samples in the low- to moderate-noise regime, which may facilitate the learning of fine-grained structures from noisy latents.
During inference, the shift instead affects the allocation of sampling steps across noise levels. Prior work suggests that global structure is largely established in the high-noise regime~\cite{ref41}. Allocating more sampling steps to this regime may therefore help reduce structural errors. However, an excessively large inference shift may leave insufficient steps for low-noise refinement, limiting the recovery of fine details.
These different considerations may explain why the preferred shifts for training and inference do not coincide. Among the tested configurations, World-R0 achieves the best generation performance with a training shift of 7.746 and an inference shift of 17.889. These results suggest that training and inference shifts can be configured separately to improve generation quality.

\begin{table}[t]
  \caption{\textbf{Joint ablation of training and inference shifts on UCF101.} Rows represent training shifts and columns represent inference shifts; values are gFVD, where lower is better. Bold indicates the lowest value.}
  \label{tab:shift-abl}
  \centering
  \begin{tabular}{lccc}
    \toprule
    \multirow{2}{*}{Training shift}
      & \multicolumn{3}{c}{Inference shift} \\
    \cmidrule(lr){2-4}
      & 7.746 & 17.889 & 35.000 \\
    \midrule
    7.746  & 115.68 & \textbf{114.96} & 116.02 \\
    17.889 & 118.52 & 116.25 & 116.92 \\
    35.000 & 124.40 & 125.13 & 125.32 \\
    \bottomrule
  \end{tabular}
\end{table}

\subsection{Action Generation}\label{sec:exp-action-generation}

\subsubsection{Experimental Setup}

We evaluate whether World-R0 representations can effectively support robotic-arm action-sequence generation on LIBERO. The experiment follows the evaluation protocol of V-JEPA-Policy and covers four task suites: Spatial, Object, Goal, and Long. Each suite contains 10 tasks with 50 trials per task, for 2,000 closed-loop trials in total. Evaluation uses random seed 42 and the EGL simulation backend. We report success rates for each suite and their average.

We freeze the World-R0 encoder and initialize the Generator and action expert from scratch. The input consists of 224$\times$224 agent-view and wrist-view images, a language instruction, and the robot's proprioceptive state. We use cached T5-XXL features with a maximum length of 128 tokens; the proprioceptive state has at most 48 dimensions. The visual encoder applies 2$\times$ temporal compression, with each tubelet covering two sampled frames. We set $\lambda_{\mathrm{future}}=1.0$.

Training proceeds in two stages. First, we use a short history containing one tubelet $W=1$ for 21,360 steps, with peak learning rate $1\times10^{-4}$. We then adapt to variable-length histories by sampling $W\in\{1,\ldots,9\}$ for 10,000 steps, with peak learning rate $3\times10^{-5}$. Both stages use global batch size 128 and AdamW with $\beta=(0.9,0.95)$ and weight decay 0.04. The first 5\% of steps use linear warmup from 0.01 times the peak rate, followed by cosine decay to zero. ``From Scratch'' in the table means that both the downstream Generator and action expert are trained from random initialization.

At inference, the future-feature horizon is 32 frames. The action expert uses 10 sampling steps and produces a 32-step action chunk with 7 dimensions per step, replanning after the first 16 steps are executed. We sample historical observations every four environment steps, with at most 9 tubelets in the window. A fixed-length sliding window and its causal KV cache are maintained, and only newly arrived observations are encoded.

\subsubsection{LIBERO Results}

As shown in Table~\ref{tab:libero}, World-R0 achieves an average success rate of 97.75\% with 0.9B parameters, outperforming the from-scratch V-JEPA-Policy baseline of comparable model size by 0.90 percentage points. Spatial, Goal, and Long each improve by 1.40 points, reaching 98.20\%, 97.20\%, and 96.60\%. Object reaches 99.00\%, which is 0.60 points below V-JEPA-Policy's 99.60\%. The improvement on LIBERO-Long demonstrates the effectiveness of World-R0 on long-horizon manipulation tasks, while its causal encoding enables incremental reuse of past computations through a sliding-window KV cache.

\begin{table}[t]
  \caption{\textbf{LIBERO action-generation results.} Success rates are percentages, and parameter counts are in billions (B). ``From Scratch'' denotes random initialization of both the downstream Generator and action expert. Avg.\ denotes the average over tasks. Bold and underline indicate the best and second-best results, respectively. * indicates reproduction using the official pretrained weights.}
  \label{tab:libero}
  \centering
  \resizebox{\textwidth}{!}{
  \begin{tabular}{lccccccc}
    \toprule
    \multirow{2}{*}{Method} & \multirow{2}{*}{\shortstack{Params.\\(B)}} & \multirow{2}{*}{Pretrain} & \multicolumn{5}{c}{LIBERO} \\
    \cmidrule(lr){4-8}
      & & & Avg. $\uparrow$ & Spatial $\uparrow$ & Object $\uparrow$ & Goal $\uparrow$ & Long $\uparrow$ \\
    \midrule
    OpenVLA-OFT \cite{openvlaoft} & 7.7 & \checkmark & 97.10 & 97.60 & 98.40 & 97.90 & 94.50 \\
    $\pi_0$ \cite{pi0} & 3.3 & \checkmark & 94.15 & 96.80 & 98.80 & 95.80 & 85.20 \\
    $\pi_{0.5}$ \cite{pi05} & 3.3 & \checkmark & 96.85 & \textbf{98.80} & 98.20 & 98.00 & 92.40 \\
    VLA-JEPA \cite{vlajepa} & 2.3 & \checkmark & 97.20 & 96.20 & 99.60 & 97.20 & 95.80 \\
    PRTS \cite{prts} & 5.0 & \checkmark & \textbf{98.40} & \textbf{98.80} & \underline{99.80} & \underline{98.40} & \underline{96.60} \\
    \midrule
    ResVLA \cite{resvla} & 4.0 & $\times$ & 96.55 & 96.00 & \textbf{100.00} & 97.40 & 92.80 \\
    StarVLA-$\pi$ \cite{starvla} & 7.8 & $\times$ & 95.65 & \textbf{98.80} & 99.60 & 95.80 & 88.40 \\
    Fast-WAM \cite{fastwam} & 6.0 & $\times$ & 97.60 & \underline{98.20} & \textbf{100.00} & 97.00 & 95.20 \\
    ImageWAM \cite{imagewam} & 4.5 & $\times$ & \textbf{98.40} & 97.20 & 99.20 & \textbf{98.80} & \textbf{98.40} \\
    JEPA-WAM \cite{jepawam} & 1.3 & $\times$ & 96.70 & 95.60 & 99.40 & 97.20 & 94.60 \\
    V-JEPA-Policy* (From Scratch) \cite{vjepa-policy} & 0.9 & $\times$ & 96.85 & 96.80 & 99.60 & 95.80 & 95.20 \\
    World-R0 (From Scratch) & 0.9 & $\times$ & \underline{97.75} & \underline{98.20} & 99.00 & 97.20 & \underline{96.60} \\
    \bottomrule
  \end{tabular}
  }
\end{table}

\subsection{3DGS Generation}

\subsubsection{Experimental Setup}

We evaluate World-R0 on single-image-conditioned 3D Gaussian scene generation using RE10K. We report gFVD, where lower values indicate a closer match between the generated and real scene distributions in the feature space. The experiment has two stages, Gaussian-decoder training followed by image-conditioned latent-DiT training, both at $256\times256$ resolution. The World-R0 encoder is frozen throughout.

In the first stage, we train the Gaussian decoder with YoNoSplat on RE10K. Training uses 16 context views and 8 target views and updates only the decoder. We run 50,000 steps with Adam at learning rate $5\times10^{-5}$, apply linear warmup for the first 500 steps, and supervise with losses for RGB MSE, LPIPS, opacity, camera-intrinsic parameters, and relative pose.

In the second stage, we train DiT separately. The input contains 15 consecutive frames, with the first frame copied once to form a 16-frame sequence. After 2$\times$ temporal compression, the sequence yields 8 latents: the first is the single-image condition and the remaining 7 are generation targets. We use only latent flow-matching loss, without RGB or 3D Gaussian supervision. DiT is trained for 100 epochs with GMuon at learning rate $2\times10^{-4}$ and 2,000 warmup steps.

\subsubsection{Generation Results}

Table~\ref{tab:3dgs-result} reports single-image-conditioned 3D Gaussian generation on RE10K. OneWorld outperforms World-R0 in terms of gFVD, achieving 83.09 compared with 93.21 for World-R0. It uses a pretrained Wan2.1 model and receives additional camera-pose inputs, benefiting from both pretrained initialization and explicit geometric conditioning. World-R0 uses only one image and its own latent, with no camera-pose input. The gap therefore reflects both sources of prior information. Even so, the 93.21 score shows that World-R0 latents contain spatial and appearance cues that a Gaussian decoder can exploit without explicit pose.

\begin{figure}[t]
  \centering
  \includegraphics[width=\textwidth]{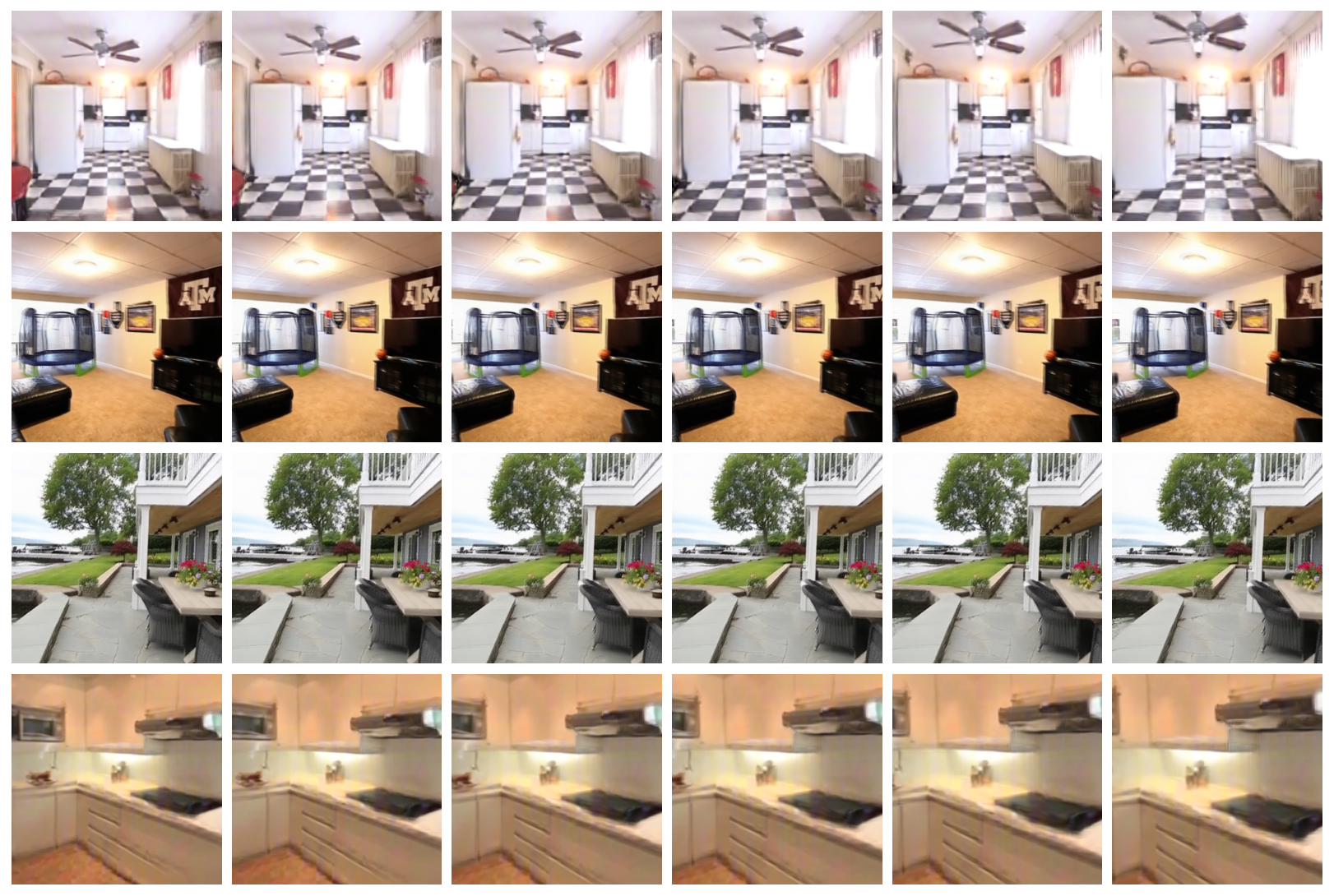}
  \caption{\textbf{Visualization of single-image-conditioned 3D Gaussian scene generation by World-R0 on RE10K.} Each row corresponds to one generated 3D scene, shown through six rendered views.}
  \label{fig:3dgs-vis}
\end{figure}

Figure~\ref{fig:3dgs-vis} shows qualitative results generated by World-R0. The room layout, positions of major objects, wall and floor structures, and overall color remain consistent across sampled frames, suggesting that World-R0 can generate a structurally coherent scene representation from a single image. Some regions remain overly smooth and locally blurred, leaving room to improve fine-detail recovery.

\begin{table}[t]
  \caption{\textbf{Single-image-conditioned 3D Gaussian scene generation results on RE10K.} Lower gFVD is better.}
  \label{tab:3dgs-result}
  \centering
  \begin{tabular}{lc}
    \toprule
    Method & gFVD $\downarrow$ \\
    \midrule
    OneWorld & \textbf{83.09} \\
    World-R0 & 93.21 \\
    \bottomrule
  \end{tabular}
\end{table}

\section{Conclusion}

We present World-R0, a general-purpose world representation that supports causal updates. To address the limitations of existing representation adaptation methods, including the lack of temporal modeling, information loss caused by bottlenecks, and degradation introduced by post-hoc adaptation, we explore a simple yet effective post-training approach based on asymmetric distillation. We train a causal encoder under the supervision of a bidirectional video teacher with full context and an image teacher that processes individual frames independently. This enables the encoder to achieve representational capabilities broadly comparable to those of bidirectional V-JEPA2.1-L while accessing only current and past frames. We also analyze feature covariance spectra and show that representations differ in their energy structure. Feature-energy-aware shift modulation uses this information to provide flow-matching generation with a better-matched noise schedule, improving video generation on all three datasets. Taken together, the results of representation probing and downstream generation experiments show that World-R0 maintains competitive visual understanding and reconstruction capabilities under causal constraints while supporting three types of world-modeling tasks through a unified representation: video generation, action generation, and 3D scene generation.

\textbf{Limitations.} The largest representation-training run uses about 2.4 million samples, so scalability to larger and more diverse data remains open. Our video generation experiments primarily use class conditioning. Due to training costs, we have not yet trained a large-scale foundation model for video generation using World-R0 representations. We have therefore not yet fully evaluated how well these representations support open-vocabulary text conditioning and controllable generation. The feature-energy-aware shift modulation is mainly derived from spectral statistics and limited ablations, and the empirical form is not claimed to be theoretically unique or optimal. Future research could expand the training scale, train a large-scale foundation model for video generation using World-R0 representations, evaluate open-vocabulary text-conditioned and controllable generation, and explore more general noise schedules.

\bibliographystyle{assets/plainnat}
\bibliography{references}

\end{document}